%% file: camera_ready.tex
\documentclass[sigconf]{acmart}  % review, anonymous

\usepackage{multirow}
\usepackage{adjustbox}
\usepackage{tabularray}
\usepackage{placeins}
\usepackage[normalem]{ulem}
\usepackage{float}

\let\oldFootnote\footnote
\newcommand\nextToken\relax

\renewcommand\footnote[1]{%
    \oldFootnote{#1}\futurelet\nextToken\isFootnote}

\newcommand\isFootnote{%
    \ifx\footnote\nextToken\textsuperscript{,}\fi}

\AtBeginDocument{%
  \providecommand\BibTeX{{%
    \normalfont B\kern-0.5em{\scshape i\kern-0.25em b}\kern-0.8em\TeX}}}

\copyrightyear{2025}
\acmYear{2025}
\setcopyright{acmlicensed}
\acmConference[MM '25] {Proceedings of the 33rd ACM International Conference on Multimedia}{October 27--31, 2025}{Dublin, Ireland.}
\acmBooktitle{Proceedings of the 33rd ACM International Conference on Multimedia (MM '25), October 27--31, 2025, Dublin, Ireland}
\acmISBN{979-8-4007-2035-2/2025/10}
\acmDOI{10.1145/3746027.3755583}
\begin{document}

% \title[Interactive Video Corpus Moment Retrieval using Reinforcement Learning]{Interactive Video Corpus Moment Retrieval \\ using Reinforcement Learning}
\title{Mitigating Cross-modal Representation Bias for Multicultural Image-to-Recipe Retrieval}

% \acmSubmissionID{2265}

\author{Qing Wang}
\email{qwang@smu.edu.sg}
\affiliation{%
  \institution{Singapore Management University}
  \city{Singapore}
  \country{Singapore}
}

\author{Chong-Wah Ngo}
\email{cwngo@smu.edu.sg}
\affiliation{%
  \institution{Singapore Management University}
  \city{Singapore}
  \country{Singapore}}

\author{Yu Cao}
\email{yu.cao.2022@msc.smu.edu.sg}
\affiliation{%
  \institution{Singapore Management University}
  \city{Singapore}
  \country{Singapore}}

\author{Ee-Peng Lim}
\email{eplim@smu.edu.sg}
\affiliation{%
  \institution{Singapore Management University}
  \city{Singapore}
  \country{Singapore}
  }

% \renewcommand{\shortauthors}{Ma and Ngo}


\input{sections/0_abstract}



% \maketitle
\maketitle

\input{sections/1_intro}

\input{sections/2_related_work}
\input{sections/3_causal_graph}

\input{sections/4_framework}
\input{sections/5_experiments}

\input{sections/6_experiments}

\input{sections/7_conclusion}
\input{sections/8_ack}
\input{sections/X_suppl}

\bibliographystyle{ACM-Reference-Format}
\bibliography{sample-base}

\end{document}

%% file: sections/0_abstract.tex
\begin{abstract}

Existing approaches for image-to-recipe retrieval have the implicit assumption that a food image can fully capture the details textually documented in its recipe. However, a food image only reflects the visual outcome of a cooked dish and not the underlying cooking process. Consequently, learning cross-modal representations to bridge the modality gap between images and recipes tends to ignore subtle, recipe-specific details that are not visually apparent but are crucial for recipe retrieval. Specifically, the representations are biased to capture the dominant visual elements, resulting in difficulty in ranking similar recipes with subtle differences in use of ingredients and cooking methods. The bias in representation learning is expected to be more severe when the training data is mixed of images and recipes sourced from different cuisines. This paper proposes a novel causal approach that predicts the culinary elements potentially overlooked in images, while explicitly injecting these elements into cross-modal representation learning to mitigate biases. Experiments are conducted on the standard monolingual Recipe1M dataset and a newly curated multilingual multicultural cuisine dataset. The results indicate that the proposed causal representation learning is capable of uncovering subtle ingredients and cooking actions and achieves impressive retrieval performance on both monolingual and multilingual multicultural datasets.

\end{abstract}

%% file: sections/1_intro.tex
\section{Introduction}
\label{sec:intro}

Cross-modal recipe retrieval offers a scalable alternative to traditional classification in food analysis~\cite{chen2017cross3, min2019survey, salvador2017learning}. Previous research~\cite{salvador2017learning, chen2018deep, zhu2019r2gan, zhu2020cookgan, salvador2021revamping} has framed recipe retrieval as a cross-modal representation learning problem, where recipes and images are encoded with separate encoders and projected into a shared embedding space to maximize pairwise similarity. An implicit assumption is that an image can capture and reflect its recipe content. Therefore, the learning aims to embed the ingredients and cooking procedure jointly observed in images and recipes into the shared space. Nevertheless, ingredients and cooking actions cannot be treated equally due to their visual impressions in food images. For example, seasoning ingredients to enhance flavor are not as visible as the major ingredients of a dish. Similarly, transformative cooking actions (e.g., cutting, baking), which change the structure and texture of ingredients, are more visible than preservative actions (e.g., salting, smoking), which only introduce subtle changes to appearance. As the visibility of ingredients and cooking actions is unequal, representation learning by maximizing the correlation between images and recipes can inevitably result in representation biases.

To address biased representation learning, we focus on removing
spurious correlations that negatively impact the accuracy of cross-modal similarity measurement. By treating both ingredients and
cooking actions as confounders in food preparation, {we propose a novel backdoor adjustment to refine cosine similarity commonly
used for this problem and thus improve retrieval performance.} Specifically, we inject two additional terms
corresponding to the representations of ingredients and cooking
actions, respectively, to reduce the biases in similarity measurement. We also propose neural networks comprising culinary-specific classifiers and dictionaries to approximate these two terms, which are lightweight and can be plugged-and-played into the existing SOTA models, including H-T~\cite{salvador2021revamping}, TFood~\cite{shukor2022transformer}, VLPCook~\cite{shukor2022structured} and multilingual CLIP variants~\cite{carlsson2022cross,visheratin2023nllb,ilharco_gabriel_2021_5143773}, for image-to-recipe retrieval.

The main contributions of this paper are twofold. First, we propose a causal view of cross-modal image-to-recipe retrieval, which leads to an elegant formulation for alleviating the representation bias. A novel backdoor adjustment is thus proposed to mitigate the representation biases introduced by ingredients and cooking methods. Second, we consider multilingual multicultural recipe retrieval, where the bias in representation can become even more severe with partially overlapping ingredients and cooking actions among cuisines. By the proposed backdoor adjustment,  we propose plug-and-play neural modules to reduce the cuisine-specific biases in representation learning. To our best knowledge, there is no prior research addressing the issue of representation bias for multicultural image-to-recipe retrieval.

%% file: sections/2_related_work.tex
\section{Related Work}
\label{sec:related_work}

Cross-modal recipe retrieval is to retrieve a recipe corresponding to a dish image or vice versa. Most approaches in this area use separate encoders to map images and recipes into a shared embedding space and maximize pairwise similarity. Recipe encoders are based on LSTMs~\cite{schmidhuber1997long}, hierarchical Transformer~\cite{salvador2021revamping}, and multilingual BERT~\cite{guerrero2021cross, zhu2022cross_mixup}. Image encoders include ResNet-50 pre-trained on ImageNet~\cite{salvador2017learning, carvalho2018cross, fu2020mcen, xie2021learning, salvador2021revamping}, and CLIP-ViT with CLIP weights~\cite{shukor2022transformer, shukor2022structured, huang2023improving}. Cross-modal multilingual alignment has seldom been explored for recipe retrieval, except for recipe augmentation~\cite{guerrero2021cross, zhu2022cross_mixup}. In X-MRS~\cite{guerrero2021cross}, multilingual BERT is exploited to augment recipes by back translation (e.g., translate an English recipe to German, and then the German recipe back to the English version) for representation learning. In Recipe Mixup~\cite{zhu2022cross_mixup}, multilingual BERT is also employed for recipe augmentation to address cross-lingual domain adaptation. None of these works address the issue of representation bias.

Recent works~\cite{shukor2022structured, huang2023improving, song2024DAR, zhao2025cross} enhance representation learning using multimodal contexts extracted from foundation models. VLPCook~\cite{shukor2022structured} utilizes CLIP to identify ingredients and titles that best match a query image as context. Similarly, FMI~\cite{zhao2025cross} uses title and ingredient features extracted from recipes to enhance image representation. Recently, DAR~\cite{song2024DAR}, employs SAM~\cite{kirillov2023segment} to segment ingredients in images. The segmented regions are used to align with Llama2-generated visual descriptions extracted from its recipe for representation learning. Rather than enriching representation with contexts as in~\cite{shukor2022structured, huang2023improving, song2024DAR, zhao2025cross},  we leverage causal inference to identify the causes and then propose backdoor adjustment to alleviate bias in representation learning. 

Causal inference has been widely applied to representation learning across various tasks~\cite{liu2022show, wang2022causal, wang2021deconfounded, lv2022causality}, focusing on single-modal image representation learning. In the context of multimodal learning,  \cite{qi2020two} identifies that the visual dialogue task is confounded by an unobserved variable (i.e., user preference), introducing spurious correlations between questions and answers. Similarly, in video moment retrieval~\cite{yang2021deconfounded}, an unobserved confounder (i.e., moment temporal location) induces spurious correlations between model input and prediction. In E-commerce cross-modal retrieval~\cite{ma2022ei}, common-sense biases learned in pretrained models are identified as confounders. Unlike these approaches, we aim to mitigate dataset bias by identifying observable confounders within the dataset.

It is worth noting that there are also efforts aimed at learning robust representations by reconstructing cooking programs~\cite{papadopoulos2022learning} and recipes~\cite{salvador2019inverse, chhikara2024fire, wang2022learning} from images. For instance, in~\cite{papadopoulos2022learning}, both food images and recipes are represented as cooking programs. To achieve this, cooking programs are first crowdsourced for the Recipe1M dataset, and a program decoder is employed to generate cooking programs based on food images or cooking recipes. By encouraging the generated programs to closely match the crowdsourced ones, improved cross-modal retrieval and food recognition performance are attained. Although these approaches are not explicitly grounded in causal theory, the multi-modal representations learned in this manner may also capture the causal effects inherent in cooking.

%% file: sections/3_causal_graph.tex
\section{Causality-based Representation Learning}
\label{sec:causal_graph}

\input{tables/causal_graph}

\textbf{A Causal View.} We denote the image, recipe, ingredient, and cooking action as $I$, $R$, $Ing$, and $Act$, respectively. Their relationships are illustrated in the directed graph in Figure~\ref{fig:causal_graph_ing_act}, with directed edges presenting the causal relationships between nodes. For example, $Ing \rightarrow R \leftarrow Act$ indicates that a recipe ($R$) is composed of (or caused by) ingredients ($Ing$) and actions ($Act$). $Ing \rightarrow I \leftarrow Act $ symbolizes an image (I) as the cause of applying a sequence of actions on ingredients as prescribed in a recipe, i.e., $R \rightarrow I$. The confounders, $Ing$ and $Act$, skew the information flow through the pathways $R \leftarrow Ing \rightarrow I$ and $R \leftarrow Act \rightarrow I$, respectively, creating spurious correlation and biased similarity measure ($S$). The distorted flow is further amplified by the fact that major ingredients and the effects of transformative cooking actions, which are more visible in food images, tend to exhibit greater influence on representation learning. The biases impede accurate cross-modal similarity measures. By Bayes rule, we model the image-recipe similarity as:
\begin{subequations} \label{eq:biasing_equation_sup}
    \begin{align}
      &  P(S|I,R) \notag \\
        & = \sum_{ing}\sum_{act}P(S,ing, act|I,R) \label{eq:existng_1_sup} \\
        & = \sum_{ing}\sum_{act}P(S|I,R, ing, act)P(ing, act|I, R) \\
        & = \sum_{ing}\sum_{act}P(S|I,R, ing, act)P(ing|I, R)P(act|I, R, ing) \label{eq:existng_2_sup}.
    \end{align}
\end{subequations}

In Eq.~(\ref{eq:existng_2_sup}), ingredients observed in both image and recipe exhibit a higher value of $P(ing|I,R)$. Conversely, ingredients, which appear in $R$ but are hardly observed in $I$, will have less impact on the similarity score, $P(S|I,R)$. Similarly, $P(act|I, R, ing)$ will bias towards cooking methods such as chopping, which will alter the visual appearance of ingredients in terms of shape and structure, more than actions such as simmering and marinating, which are harder to observe in the image. Note that the subtle variations introduced by ingredients and cooking methods, which are poorly or partially captured in images, play a crucial role in $P(S|I,R)$ for disambiguating similar recipes. This includes recipes that use the same ingredients but differ in cooking methods, as well as those that share both cooking technique and ingredients but vary in seasoning.

\textbf{Backdoor Adjustment.} To remove spurious correlations caused by the confounders, we apply backdoor adjustment to intervene the image variable (i.e., $do(I)$) by removing all the incoming edges to image $I$ (Figure~\ref{fig:causal_graph_ing_act} (right)), resulting in the similarity measure as:
     \begin{subequations} \label{eq:debiasing_equation}
    \begin{align}
   & P(S|do(I),R) \notag \\
   & = \sum_{ing}\sum_{act} P(S|do(I),R,ing, act){P(ing, act|do(I),R)} \label{eq:dev_act1} \\ 
       & = \sum_{ing}\sum_{act} P(S|do(I),R,ing, act){P(ing, act|R)} \label{eq:dev_act2} \\ 
       & = \sum_{ing}\sum_{act} P(S|I,R,ing, act){P(ing,act|R)} \label{eq:dev_act3} \\
       & = \sum_{ing}\sum_{act} P(S|I,R,ing, act){P(ing|R)}{P(act|R, ing)}, \label{eq:dev_act4}
    \end{align}
\end{subequations}
where by the rule-3 of $do$-calculus (Theorem 3.4.1~\cite{pearl2009causality}), the $do(I)$ in $P(ing, act|do(I),R)$ can be omitted. This is due to $S$ is a collider of $R$ and $I$ and blocks the information flow from $Ing$ to $I$, i.e.,  $Ing \rightarrow R \rightarrow S \leftarrow I$. Hence, $P(ing, act|do(I),R)=P(ing, act|R)$ and we have Eq.~(\ref{eq:dev_act2}). By rule-2 of do-calculus, we obtain Eq.~(\ref{eq:dev_act3})  because $S$ is independent of $I$ after removing the outgoing edges from $I$. Using the chain rule of conditional probability, we decompose $P(ing, act|R)$ in Eq.~(\ref{eq:dev_act3}) and arrive at Eq.~(\ref{eq:dev_act4}).

\input{tables/action_framework}

\textbf{Neural Approximation.} Eq.~(\ref{eq:dev_act4}) mitigates the bias by weighting the similarity with the true distributions of ingredients and actions in a recipe, denoted as $P(ing|R)$ and $P(act|R, ing)$, rather than the confounded distributions $P(ing|I, R)$ and $P(act|I, R, ing)$. In Eq.~(\ref{eq:dev_act4}), we set $P(S|I, R, ing, act) = f_s(e_I,e_R, e_{ing}, e_{act})$, where $f_s()$ is a similarity function, and $e_I, e_R, e_{ing}, e_{act}$ are the embedding of image $I$, recipe $R$, ingredient $Ing$ and action $Act$, respectively:
\begin{subequations} \label{eq:implement_equation1}
    \begin{align}
    & P(S|do(I),R) \notag \\
    & = \sum_{ing}\sum_{act} f_s(e_I,e_R, e_{ing}, e_{act})P(act|R, ing)P(ing|R) \label{eq:implement_eq1_act} \\
&\approx e_R \cdot \Bigg(e_I + \sum_{ing}P(ing|I) \cdot e_{ing} \notag \\
    &\quad + P(ing_1|I) \cdot \sum_{act}P(act|I, ing_1)\cdot e_{act} \notag \\
    &\quad + \ldots + P(ing_K|I) \cdot  \sum_{act}P(act|I, ing_K)\cdot e_{act} \Bigg) \label{eq:eq3_1_act_expand_expectation1},
    \end{align}
\end{subequations}
where Eq.~(\ref{eq:eq3_1_act_expand_expectation1}) approximates the backdoor adjustment formula. Please refer to {Section E} of the supplementary document for the full derivation. In Eq.~(\ref{eq:eq3_1_act_expand_expectation1}), besides estimating $P(Ing_i|I)$, the actions associated with an ingredient $\sum_{act}P(act|I, Ing_i)$ are also estimated.

\textbf{Discussion}\label{sec:discus}. To facilitate the comparison between Eq.~(\ref{eq:eq3_1_act_expand_expectation1}) and the conventional similarity measure in~\cite{salvador2021revamping}, we simplify Eq.~(\ref{eq:eq3_1_act_expand_expectation1}) to:\begin{subequations} \label{eq:implement_equation2}
    \begin{align}
    & P(S|do(I),R) \notag \\ 
    & \approx e_R \cdot \Bigg( e_I + \underbrace{ \sum_{ing} P(ing|I) \cdot e_{ing} }_{\text{ingredient debiasing}} + \underbrace{ \sum_{act} P(act|I, \hat{Ing}) \cdot e_{act}}_{\text{action debiasing}} \Bigg)\label{eq:implement_eq5_act} \\
    & = e_R \cdot (e_I + e_{Ing} + e_{Act})\label{eq:implement_eq5_act_vsimple}.
    \end{align}
\end{subequations}

The term $\sum_{k}p(ing_k|I)\sum_{act}p(act|I, ing_k) \cdot e_{act}$ in Eq.~(\ref{eq:eq3_1_act_expand_expectation1}) is abbreviated as $e_{Act}=\sum_{act} P(act|I, \hat{Ing}) \cdot e_{act}$, where $\hat{Ing}$ represents ingredient composition which is introduced to simplify the visualization of the equation. Eq.~(\ref{eq:implement_eq5_act_vsimple}) extends the conventional dot product term~\cite{salvador2021revamping} (i.e., $e_R \cdot e_I$) with two debiasing terms. The first term adjusts image representation $e_I$ by adding a linear sum of ingredient embeddings weighted by their probabilities. The second term performs adjustment by supplementing $e_I$ with cooking action embeddings conditioned on the ingredient composition.

%% file: tables/causal_graph.tex
\begin{figure}
    \centering
    \includegraphics[width=0.7\columnwidth]{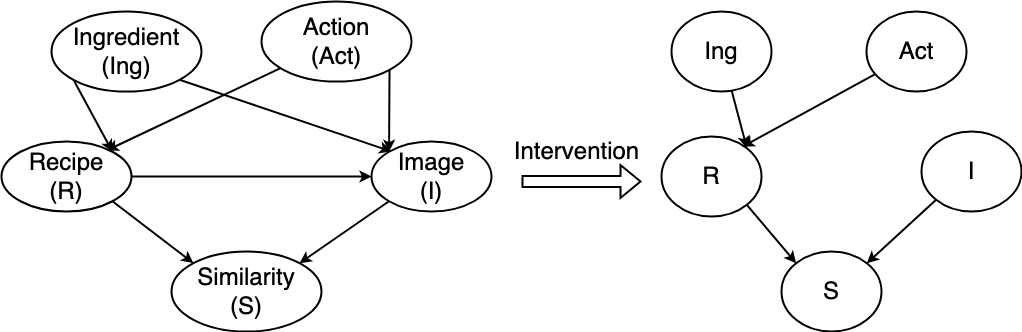}
    \caption{Left: Causal graph with ingredients and actions as confounders. Right: Backdoor adjustment mitigates spurious correlations by removing incoming edges to the image node.}
    \label{fig:causal_graph_ing_act}
\end{figure}

%% file: tables/action_framework.tex
\begin{figure*}[h!]
    \centering
    \includegraphics[width=0.8\linewidth]{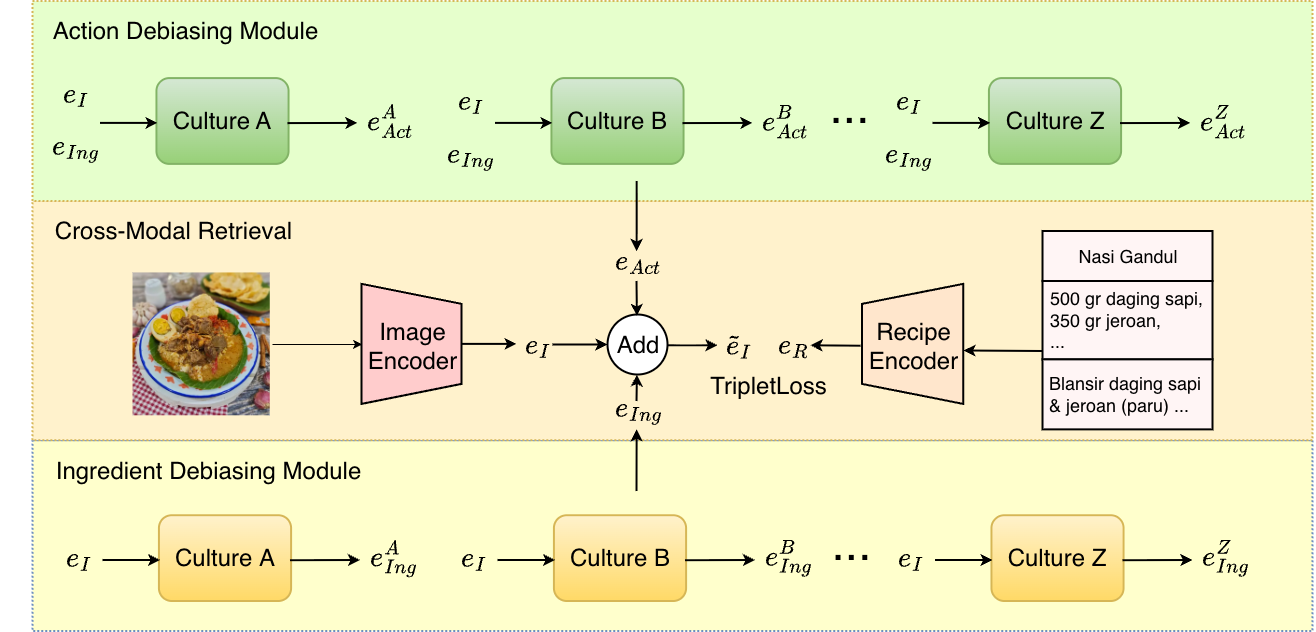}
    \caption{The proposed framework for multicultural recipe retrieval. Given a query image of Culture X, the ingredient and action debiasing modules in the culture will derive the embeddings $e_{Act}$ and $e_{Ing}$, respectively. The two embeddings are then added to the image embedding, $e_I$, learnt globally in the cross-modal retrieval module for alleviating representation biases. Please refer to Figure~\ref{fig:subfig_ingre_debiasing} and Figure~\ref{fig:subfig_action_debiasing} for the architectures of ingredient debiaisng module and action debiasing module, respectively.}
    \label{fig:framework_action}
\end{figure*}

%% file: sections/4_framework.tex
\section{Multi-lingual Multi-cultural Recipe retrieval}\label{sec:framework}

The two debiasing terms in Eq.~(\ref{eq:eq3_1_act_expand_expectation1}) can be implemented using one neural network to predict the presence of ingredients conditioned on an image, and another network to predict the presence of cooking actions conditioned on the predicted ingredients. These two networks can be "added" or plugged into the existing cross-modal representation models~\cite{salvador2021revamping, shukor2022transformer, shukor2022structured} to alleviate the potential bias in representation learning. The existing models, nevertheless, consider mostly retrieving recipes from a dataset composed of monolingual Western-dominated cuisines (e.g., Recipe1M~\cite{salvador2017learning}). In this paper, we further explore the proposed work for multilingual multicultural recipe retrieval. Specifically, in a multi-cuisine dataset, the recipes are written in different native languages. Two cuisines can differ in terms of ingredient and cooking action distributions, and only share a partial set of ingredients and cooking techniques. Learning to remove representation bias in such a scenario is highly challenging.

Figure~\ref{fig:framework_action} depicts the overall framework, where the cross-modal retrieval module is plugged with culture-specific ingredients and action debiasing modules based on Eq.~(\ref{eq:eq3_1_act_expand_expectation1}). The ingredient debiasing module (Figure~\ref{fig:subfig_ingre_debiasing}) predicts ingredient distribution using a multi-label classifier, then retrieves relevant ingredients from an ingredient dictionary. Meanwhile, the action debiasing (Figure~\ref{fig:subfig_action_debiasing}) module generates a sequence of cooking actions with a generation model, followed by retrieving corresponding actions from an action dictionary. The dual modules are specifically tailored and trained for each culture. In other words, each culture maintains its own local predictors and dictionaries to debias the image representations globally learned in the cross-modal retrieval module.

\textbf{Cross-modal retrieval}. The image encoder can be implemented with ResNet-50~\cite{he2016deep} or Vision Transformer (ViT)~\cite{dosovitskiy2020image}. In a similar way, the recipe encoder can be implemented with a hierarchical transformer~\cite{salvador2021revamping} to embed the three sections (i.e., title, ingredients, and cooking instructions) of a recipe. We employ multilingual CLIP variants~\cite{carlsson2022cross, visheratin2023nllb, ilharco_gabriel_2021_5143773} for embedding both images and recipes written in different native languages to derive image embedding $e_I$ and recipe embedding $e_R$. We finetune the CLIP models using the recipes of all cultures in a dataset. 

\textbf{Culture-specific dictionary}. Training a universal dictionary comprising culinary elements of different cultures is not practical. In general, the usage and popularity of ingredients and culinary techniques vary across cultures. For example, in Vietnam, ingredients such as ``rice paper'' are unique and almost never used in other regions such as Indonesia, Malaysia, or India. Similarly, cooking actions such as ``tempering'' are popular in India but rarely used in Indonesia, Malaysia, or Vietnam. Hence, we propose culture-specific dictionaries for debiasing. For each culture, we compile the most frequent ingredients and actions from the training recipes, and store their embeddings in the respective dictionaries. The embeddings are encoded by the recipe encoder of the cross-modal retrieval module. During training, the embeddings stored in dictionaries are frozen while the recipe decoder is finetuned.

\input{tables/subfig_ingre_debiasing}
\textbf{Ingredient debiasing module}, which aims to implement $e_{Ing}=\sum_{ing}P(ing|I) \cdot e_{ing}$ in Eq~(\ref{eq:implement_eq5_act}), uses a multi-label classifier~\cite{liu2021query2label} to predict the ingredient probability distribution, as shown in Figure~\ref{fig:subfig_ingre_debiasing}. Specifically, we employ a Transformer decoder as the classifier, feeding image embeddings $e_I$ as both key and value, while using learnable label embeddings as queries. Using a sigmoid function, only ingredients with a probability above 0.5 are used for debiasing. The probabilities of selected ingredients are normalized to sum to 1. The ingredient embeddings are then retrieved from the dictionary and linearly combined, weighted by their probabilities, as shown in Figure~\ref{fig:subfig_ingre_debiasing}. Note that the selected ingredients $e_{Ing_{k}}$ will be channeled to the action debiasing module for further processing.

\input{tables/subfig_action_debiasing}
\textbf{Action debiasing module}, implements the second term $e_{Act}$ in Eq~(\ref{eq:implement_eq5_act}), and employs the action decoder~\cite{salvador2019inverse}, as shown in Figure~\ref{fig:subfig_action_debiasing}. Conditioned on the image embedding $e_I$ and a predicted ingredient embedding $e_{Ing_k}$, the decoder generates cooking action sequences. An action embedding of an ingredient is computed by retrieving the corresponding action embeddings from the action dictionary, weighted by the normalized action probabilities. After computing the action embeddings for each ingredient ($e^1_{Act}, e^2_{Act}, ..., $), we obtain the final action embedding $e_{Act}$ by normalizing the ingredient probabilities and calculating the weighted sum of these embeddings associated with each ingredient using the normalized probabilities. The final action embedding $e_{Act}$ is then used to adjust the image representation, aligning it with the recipe embeddings.

\textbf{Training Objective.} The overall training loss combines a bi-directional triplet loss, ${\mathcal L_{triple}}$~\cite{wang2019learning}, to bring image-recipe pairs closer in the joint embedding space; a classification loss, $\mathcal{L}_{cls}$~\cite{Ridnik_2021_ICCV}, for training the ingredient classifier; and a generation loss, $\mathcal{L}_{gen}$, for the action generator.

For the multi-label ingredient classifier, we adopt the asymmetric loss~\cite{Ridnik_2021_ICCV} to address the challenges of long-tailed distribution of ingredients. Given the ingredient probabilities $p = [ p_{ing_1}, \ldots, p_{ing_k}]$ for an image $I$, the loss function for $I$ is defined as: 
\begin{equation}
    \mathcal{L}_{I}=\frac{1}{K} \sum_{k=1}^K \begin{cases}\left(1-p_{ing_k}\right)^{\gamma+} \log \left(p_{ing_k}\right), & y_{ing_k}=1, \\ \left(p_{ing_k}\right)^{\gamma-} \log \left(1-p_{ing_k}\right), & y_{ing_k}=0,\end{cases}
    \label{eq:asymmetric_loss}
\end{equation}
where $y_{ing_k}$ indicates the presence of the ingredient. Parameters $\gamma+$ and $\gamma-$ adjust the weighting for positive and negative samples, set empirically to $\gamma+ = 1$ and $\gamma- = 1$.

For the action debiasing module, $\mathcal{L}_{gen}$ is implemented using cross-entropy loss:
\begin{equation}
    \mathcal{L}_{\text{gen}}= -\frac{1}{L} \sum\limits_{l=1}^L \sum\limits_{t=1}^T \log p_\theta(y_t^l \mid y_{1:t-1}^l),
\end{equation}
where $y_t^l$ represents the probability of the $t^{th}$ action for $l^{th}$ generated ingredient and $L$ is the number of generated ingredients. The overall objective function is defined as:
\begin{equation}  \label{eq:total_loss_with_action}
    \mathcal{L} = \mathcal{L}_{triple} + \lambda_{cls} \mathcal{L}_{cls} + \lambda_{gen} \mathcal{L}_{gen},
\end{equation}
where $\lambda_{cls}=0.001$ and $\lambda_{gen}=0.001$ are hyperparameters to balance the triple loss and two debiasing losses. 

\textbf{Training Procedure.} For monolingual recipe retrieval, we train the framework in Figure~\ref{fig:framework_action} in three steps. First, the cross-modal retrieval models, specifically the image and recipe encoders, are fine-tuned from the pretrained weights. Second, leveraging the encoders, the ingredient and action dictionaries are constructed. Finally, the three modules (retrieval model, ingredient classifier, and action generator) are trained end-to-end, while the ingredient and action embeddings in the dictionaries are frozen without further updating. For multilingual recipe retrieval, we leverage multilingual CLIP variants that have been pretrained on billions of image-text pairs. These pretrained models are directly used to extract ingredient and action embeddings for dictionary construction. Subsequently, the two debiasing modules are plugged into the cross-modal retrieval module, with multilingual CLIP variants as the image and recipe encoders, for end-to-end training.

%% file: tables/subfig_ingre_debiasing.tex
\begin{figure}[h!]
    \centering
    \includegraphics[width=0.95\linewidth]{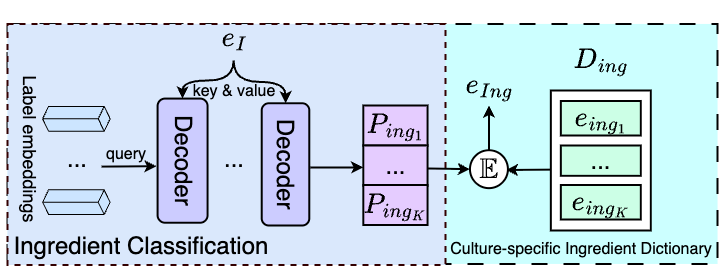}
    \caption{The ingredient debiasing module takes image embeddings $e_I$ act as keys and values, and ingredient label embeddings as queries. The decoder output is passed to a sigmoid to produce ingredient probabilities $P_{ing}$, which weight the dictionary $D_{ing}$ to yield the debiasing embedding $e_{Ing}$.}
    \label{fig:subfig_ingre_debiasing}
\end{figure}

%% file: tables/subfig_action_debiasing.tex
\begin{figure}[h!]
    \centering
    \includegraphics[width=0.95\linewidth]{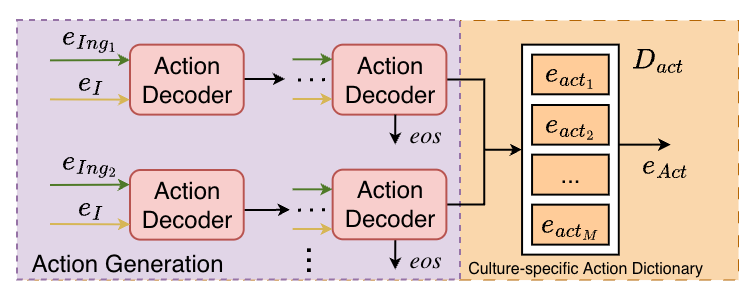}
    \caption{The action debiasing takes the predicted ingredients as input. For each ingredient $e_{Ing_k}$, we generate the sequence of cooking actions, and then retrieve the corresponding action embeddings from the dictionary $D_{act}$. We normalize the action prediction probabilities to weight each action embedding, and then compute a weighted sum of the embeddings to obtain the action embedding $e^k_{Act}$. The final action embedding, $e_{Act}$, used to enhance the image representation, is obtained by first normalizing the ingredient probabilities and then using probabilities to compute a weighted sum of the action embeddings, $e^k_{Act}$, corresponding to each ingredient. Decoders are shared by all ingredients for generation.}
    \label{fig:subfig_action_debiasing}
\end{figure}

%% file: sections/5_experiments.tex
\section{Experiment I: Monolingual Recipe Retrieval}
\label{sec:experiment1}

To validate the proposed backdoor adjustment, we conduct experiments on a monolingual recipe dataset, Recipe1M~\cite{salvador2017learning}. The dataset contains 238,999, 51,119, and 51,303 image-recipe pairs for training, validation, and testing, respectively. We sample search sets in multiples of 10K, and unless otherwise specified, we follow the evaluation protocol~\cite{salvador2021revamping} to report performance on 1K and 10K test sets. For each test set, we conduct 10 random samplings and report the average performance. The evaluation metrics are median rank (medR) and Recall@K, where K={1,5,10}. For retrieval models, lower medR and higher Recall@K indicate better retrieval performance. 

\textbf{Implementation details} We follow the settings of baseline methods, i.e., H-T~\cite{salvador2021revamping}, TFood~\cite{shukor2022transformer}, VLPCook~\cite{shukor2022structured}. For image encoders, we adopt ResNet-50, ViT-B/16, and CLIP-ViT-B/16 for H-T, TFood, and VLPCook, respectively, where CLIP-ViT-B/16 is initialized with CLIP weights while the rest two with ImageNet weights. For recipe encoders, we use transformer encoders with 2 layers and 4 heads for all three models. For the multi-label ingredient recognition, similar to~\cite{liu2021query2label}, 1 Transformer encoder layer and 2 Transformer decoder layers are utilized and both have 4 heads. For the action decoder, following~\cite{salvador2019inverse}, we employ a transformer with 4 blocks and 2 multi-head attention. The batch size is 64 and Adam optimizer is used with a base learning rate $10^{-4}$ for H-T and $10^{-5}$ for the rest. The ingredient and action debiasing models contain approximately 75M and 65M parameters, respectively.

\textbf{Model zoo} As discussed in Sec.~\ref{sec:discus}, Eq.~(\ref{eq:implement_eq5_act_vsimple}) offers three distinct approaches for debiasing retrieval models:

\begin{itemize}
    \item \textbf{+Ingredient}: ingredient-only debiasing (i.e., $e_R \cdot e_I + e_R \cdot e_{Ing}$), where a multi-label ingredient classifier predicts the ingredient distribution, and corresponding ingredient embeddings are retrieved to augment the image embeddings.
    \item \textbf{+Action}: action-only debiasing (i.e., $e_R \cdot e_I + e_R \cdot e_{Act}$). An action generator predicts the action distribution, which is used to enhance the image embeddings.
    \item \textbf{+Both}: debiasing both ingredients and actions (i.e., $e_R \cdot e_I + e_R \cdot e_{Ing} + e_R \cdot e_{Act}$), where a multi-label ingredient classifier and a conditional action generator are employed to refine the image embeddings, as shown in Figure~\ref{fig:framework_action}. 
\end{itemize}

\subsection{Performance Comparison}

Table~\ref{tab:image2recipe_split_10k_act} shows the results of image-to-recipe retrieval. Debiasing the model with ingredients or actions leads to the same medR value across the different sizes of the test set. Meanwhile, debiasing ingredients introduces more degree of improvement over cooking actions in terms of Recall@1 with around $1\%$ Recall@1 difference. Debiasing both ingredients and actions yields the best retrieval performance, improving the medR of baselines (H-T, VLPCook) by 1.0 on 10K test set. A consistent improvement is also observed, ranging from 1.7\% to 5.6\% of difference in Recall@1 across different test sizes. Compared to the most recently published results in DAR~\cite{song2024DAR}  and FMI~\cite{zhao2025cross}, our results achieves better performance in terms of R@1, R@5, and R@10 on 10K test set.

\input{tables/sota_comparison_10k_act}

We attribute the improvement over the ingredient-only or action-only debiasing to the ability to distinguish the recipes sharing similar sets of ingredients or actions. Figure~\ref{fig:ret2action_analysis_1} shows an example where H-T+ingredient cannot distinguish ``Carrot pineapple cupcakes" and ``Nutneg cookies logs", which share a similar set of ingredients (i.e., eggs, butter, white sugar, and flour). However, by predicting the actions (i.e., grease and insert) that are unique to the query image, and augmenting both the predicted ingredients and actions to the image embedding, the ground-truth recipe is alleviated from $55^{th}$ (by H-T ingredient) to the top-1 position. Note that using H-T+action alone cannot distinguish these two recipes due to some shared cooking actions (e.g., preheat, mix). More results and examples, including recipe-to-image retrieval, can be found in the supplementary document.

\input{tables/action_qualitative}

\subsection{Robustness Test}

\input{tables/scalability_test}

\textbf{Scalability.} In this section, we present the retrieval performance on larger test set sizes ranging from 20K to 50K for image-to-recipe, as shown in Table~\ref{tab:scalability}. We can observe debiasing with either ingredients or actions yields consistent improvements as test set sizes increase, though the ingredient-only module achieves slightly better results. However, the best results are achieved by debiasing both ingredients and actions, leading to additional gains in Recall@1 ranging from 0.8\% to 3.8\% across all test sizes.

\textbf{Zero-Shot Retrieval.} We evaluate the model's robustness in retrieving unseen food categories, i.e., zero-shot retrieval. To do so, we exclude all recipes from specific categories in both the training and validation sets. For instance, recipes from any category containing the word burger (e.g., turkey burgers, chicken burgers) are removed. In total, 78 dish categories, involving 14,415 recipes, are excluded. These categories are further grouped based on the removed keywords, and the search results are presented in Table~\ref{tab:zero_shot_food}. For example, the first row labeled ``pizza" shows the median rank (medR) results for all queries categorized as ``pizza''-related in the test set. Debiasing H-T with ingredients yields substantial improvements in medR across all categories. With the addition of the action debiasing module, some categories, such as ``pizza'' and ``cheesecake'', show improved medR performance. For instance, in the case of ``pizza'', the H-T model with ingredient debiasing often confuses it with visually similar dishes like ``frittatas'' and ``pies'', which are seen during training. However, the action debiasing module correctly generates actions such as preheat, bake, and spread, which are relatively unique to the pizza-making process, allowing the model to rank pizzas higher. However, if the primary cooking styles are misidentified, the action debiasing module can degrade retrieval results. For instance, while the major ingredients for a ``burger'' dish, such as beef or chicken, can be identified, the module might generate actions like spreading and topping instead of the key cooking methods for burgers, such as grilling or baking. This creates confusion with sandwich dishes, deteriorating the rank of the sandwich dish and resulting in a higher medR value for burgers.

\input{tables/zero_shot}

%% file: tables/sota_comparison_10k_act.tex
\begin{table}
\centering
\caption{Comparison on 1k and 10k test sets for image-to-recipe retrieval. medR ($\downarrow$), Recall@k ($\uparrow$) are reported. The proposed debiasing boosts the performance of existing cross-modal retrieval methods (H-T, TFood, VLPCook), especially on the 10k set. } 
\resizebox{0.95\columnwidth}{!}{
\begin{tabular}{@{}l|cccc|cccc@{}}
\toprule
\multirow{2}{*}{} & \multicolumn{4}{c|}{\textbf{1k}} & \multicolumn{4}{c}{\textbf{10k}} \\ \cmidrule(l){2-9} 
                  & medR     & R@1      & R@5    & R@10    & medR     & R@1      & R@5    & R@10   \\ \midrule
% RIVAE~\cite{kim2021learning}  & 2.0      & 39.0    & 70.0      &   79.0     &   -       & -       &     -   &       -  \\
% R2GAN \cite{zhu2019r2gan}  & 2.0     & 39.1    & 71.0    & 81.7      &  13.9        &   13.5      &   33.5    &     44.9   \\
% MCEN \cite{fu2020mcen}  & 2.0       & 48.2      &    75.8   &      83.6  &  7.2        &      20.3   &    43.3   &   54.4     \\
% ACME \cite{wang2019learning}  & 1.0      &   51.8      &   80.2    &    87.5    &     6.7     &   22.9     &  46.8     &    57.9    \\
% SN \cite{zan2020sentence}  & 1.0      & 52.7        &     81.7  &    88.9    &    7.0      &    22.1    &    45.9   &    56.9    \\
% IMHF \cite{li2021cross}  & 1.0      & 59.4        &     81.0  &    87.4    &    3.5      &    36.0    &    56.1   &    64.4    \\
% SCAN \cite{wang2021cross}  & 1.0      & 54.0        &     81.7  &    88.8    &    5.9      &    23.7    &    49.3   &    60.6    \\
% HF-ICMA \cite{li2021hybrid}  & 1.0      & 55.1        &     86.7  &    92.4    &    5.0      &    24.0    &    51.6   &    65.4    \\
% MSJE \cite{xiez2021learning}  & 1.0      & 56.5        &     84.7  &    90.9    &    5.0      &    25.6    &    52.1   &    63.8    \\
% SEJE \cite{xie2021learning}  & 1.0      & 58.1        &     85.8  &    92.2    &    4.2      &    26.9    &    54.0   &    65.6    \\
% M-SIA \cite{li2021multi}  & 1.0       &     59.3    &    86.3   &    92.6    &     4.0    &    29.2     &    55.0   &    66.2    \\
% RDE-GAN \cite{sugiyama2021cross}  & 1.0      & 55.1        &     86.7  &    92.4    &    5.0      &    24.0    &    51.6   &    65.4    \\
X-MRS \cite{guerrero2021cross}  & 1.0       &     64.0    &    88.3   &    92.6    &     3.0    &    32.9     &    60.6   &    71.2    \\
FARM~\cite{wahed2024fine}  &   1.0    &  73.7   &   90.7    &    93.4     &    2.0   &      44.9  &     71.8   &    80.0    \\
CREAMY~\cite{zou2024creamy}  &   1.0    &  73.3   &   92.5    &    95.6     &    2.0   &      44.6  &     71.6   &    80.4    \\
CIP~\cite{huang2023improving}  &    1.0     &   77.1   & 94.2   &   97.2     &  2.0     &   44.9    &   72.8  &  82.0    \\
DAR~\cite{song2024DAR}  &   1.0    &   77.3  &   95.3    &   97.7      &    2.0   &      47.8  &     75.9   &    84.3    \\ 
FMI~\cite{zhao2025cross}  &    1.0   &  77.4   &    95.8   &   97.6     &     1.0     &   48.4     &   76.3     &      81.9   \\ \midrule
H-T~\cite{salvador2021revamping}  &  1.0  &  61.8 &   88.0  & 93.2  &  4.0  &  29.9 & 58.3  &  69.6  \\
\textbf{+Ingredient}  &  1.0  & 65.7  &  89.8   & 94.1 &   3.0   &  34.4 &  62.9 &  73.6 \\ 
\textbf{+Action}  &  1.0  &  63.6 &  88.1  & 92.6 &  3.0    & 32.1  & 60.3 & 71.1  \\ 
\textbf{+Both}  &  1.0  & 65.7  &  88.8   & 93.6 &  3.0  & 35.5  &  63.8 &  74.1 \\ \midrule

TFood~\cite{shukor2022transformer}  &  1.0  &  72.4 &   92.5  & 95.4 &   2.0   & 43.9  & 71.7  & 80.8  \\
\textbf{+Ingredient}  &  1.0  &  74.5   & 93.2 & 96.1  & 2.0  & 45.6  &  73.0 & 81.6  \\ 
\textbf{+Action}  &  1.0  & 73.8  &  93.1  & 95.8 &   2.0   & 45.1  & 72.6 &  81.3 \\ 
\textbf{+Both}  &  1.0  &  75.8 &  93.6   & 96.3 &  2.0  &  46.9 &  74.4 & 82.8  \\ \midrule
VLPCook~\cite{shukor2022structured}  &  1.0  &  77.4 &   94.8  & 97.1  &   2.0   & 48.8  & 76.2  & 84.5  \\
\textbf{+Ingredient}  &  1.0  & 78.3  &  95.1   & 97.4  &  1.4    & 50.2  & 77.3 & 85.2  \\ 
\textbf{+Action}  &  1.0  & 77.9  &  95.0  & 97.4 &  1.5 &  50.0 &  77.4 &  85.4 \\ 

\textbf{+Both}  &  1.0  & 79.1  &  94.6   & 97.0 &  1.0  & 51.7  & 78.2  &  85.9 \\ % \midrule
% Oracle  &  1.0  &  99.0 &  99.8   & 99.9  &  1.0    & 96.2 & 99.2  &  99.5  \\
\bottomrule
\end{tabular}
}
\label{tab:image2recipe_split_10k_act}
\end{table}

%% file: tables/action_qualitative.tex
\begin{figure}
    \centering
    \includegraphics[width=0.98\linewidth]{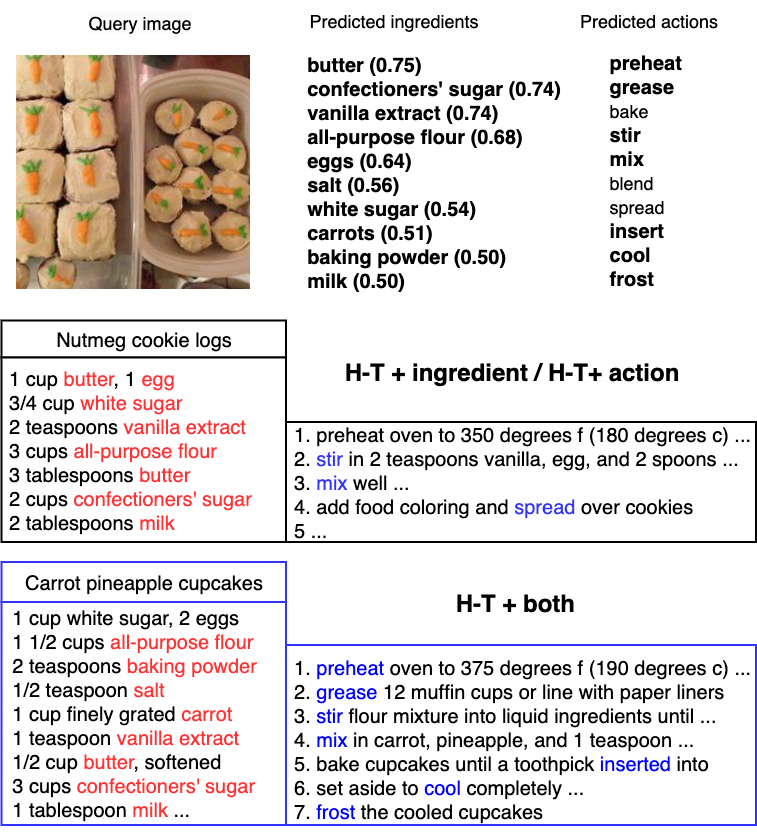}
    
    \caption{Example showing how the ingredient and action debiasing disambiguates similar recipes. The first row displays the query image, predicted ingredients, and predicted actions. The second row is the retrieved recipe by H-T+ingredient and H-T+action, while the last row is the recipe retrieved by H-T+both. The correctly predicted ingredients and cooking actions are bolded. The predicted ingredients and cooking actions are marked in red and blue, respectively, in the recipes. The ground-truth is boxed in blue.}
    \label{fig:ret2action_analysis_1}
\end{figure}

%% file: tables/scalability_test.tex
\begin{table}[]
    \centering
    \caption{Scalability test on 20k, 30k, 40k and 50k test set.}
    \begin{adjustbox}{width=0.9\columnwidth}
    \begin{tabular}{@{}l|cc|cc|cc|cc@{}}
\toprule
\multirow{2}{*}{} %& \multicolumn{8}{c|}{\textbf{1k}}                                                    & \multicolumn{8}{c}{\textbf{10k}} \\ \cmidrule(l){2-17} 
                  & \multicolumn{2}{c|}{\textbf{20k}} & \multicolumn{2}{c|}{\textbf{30k}} & \multicolumn{2}{c|}{\textbf{40k}} & \multicolumn{2}{c}{\textbf{50k}} \\ 
                  & medR     & R@1      & medR     & R@1       & medR     & R@1        & medR     & R@1       \\ \midrule

H-T~\cite{salvador2021revamping}            &  6.3  &  22.2  & 9.0 &  18.4      &  12.0    & 16.0   & 15.0 & 14.3   \\ 
\textbf{+Ingredient}           &  5.0  & 26.2   & 7.0 &  22.0       &   9.0   & 19.3   & 11.0 & 17.4        \\ 
\textbf{+Action}           &  5.8  & 24.3   & 8.0 &  20.3       &  10.0    &  17.8  &  12.6  &  15.9       \\ 
\textbf{+Both}           &  4.7  &  27.3  & 6.0 &   23.0      &   8.0   &  20.3  &  10.0  &  18.2        \\ \midrule

TFood~\cite{shukor2022transformer}           &  3.0  & 35.5  &  4.0 &  30.9      &   5.0   & 27.8 & 6.0  & 25.7       \\ 
\textbf{+Ingredient}            &  3.0  & 37.6   & 3.0 &  32.9      &   4.0   & 29.9   & 5.0  & 26.9          \\ 
\textbf{+Action}           & 3.0   &  36.3  & 4.0 &   31.6      &    4.0  &  28.5  &  5.0  &  26.2       \\ 
\textbf{+Both}           &  2.0  &  38.6  & 3.0 &   33.6      &  4.0    &  30.4  &  5.0  &    28.1      \\ \midrule

VLPCook~\cite{shukor2022structured}          &  2.0  &  40.2  & 3.0 &   35.2    &   4.0   & 32.0  &  4.0 & 29.7                   \\ 
\textbf{+Ingredient}            &  2.0  & 41.7    &  3.0  &  36.9   & 3.0 & 33.7   & 4.0  & 31.1       \\ 
\textbf{+Action}           &  2.0  &  41.0  & 3.0 &  36.0       &   3.0   &  32.7  &  4.0  &  30.2       \\ 
\textbf{+Both}           &  2.0  & 42.7   & 3.0 &  37.7       &  3.0    &  34.5  &  4.0  &  32.0        \\ %\midrule

% Oracle           &  1.0  & 94.6  & 1.0 & 93.5     &  1.0    & 92.6   & 1.0 & 91.8                    \\ 
\bottomrule
\end{tabular}
    \end{adjustbox}
\label{tab:scalability}
\end{table}

%% file: tables/zero_shot.tex
\begin{table}[]
    \centering
    \caption{Median rank comparison for unseen dish categories on the 50k test set.}
    \begin{adjustbox}{width=0.7\linewidth}
     \begin{tblr}{cccccccc}
\hline
Food type   & Oracle  & H-T &  {H-T\textbf{+Ingredient}} & {H-T\textbf{+Both}} \\ \hline
pizza & 1.0 & 23.0 & 20.0 & 16.0 \\ \hline 
steak & 1.0 & 27.0 & 19.0   & 18.0  \\ \hline 
pancakes & 1.0 & 32.0 & 19.0  & 19.0   \\ \hline 
cheesecake & 1.0 & 29.0 & 18.0  & 16.0  \\ \hline 
cupcake & 1.0 & 22.0 & 19.0   & 12.0   \\ \hline 
lasagna & 1.0 & 18.0 & 12.0  & 15.0  \\ \hline 
rice & 1.0 & 15.0 & 11.0  & 12.0    \\ \hline 
tacos & 1.0 & 17.0 & 11.0  & 12.0    \\ \hline 
burger & 1.0 & 23.0 & 11.0 & 17.0  \\ \hline 
waffles & 1.0 & 19.0 & 12.0  & 12.5   \\ \hline 

\end{tblr}
\end{adjustbox}
        \label{tab:zero_shot_food}
\end{table}

%% file: sections/6_experiments.tex
\section{Experiment II: Multicultural Recipe Retrieval}
\label{sec:experiment2}

\input{tables/cookpad_statistics}

\subsection{Dataset Curation}
Next, we conduct the experiments on a newly curated dataset composed of five different cultures: Indonesia, Malaysia, Thailand, Vietnam, and India. The image-recipe pairs are crawled from Cookpad\footnote{https://cookpad.com/}, using the dish titles compiled from Wikipedia\footnote{https://en.wikipedia.org/wiki/List\_of\_Indonesian\_dishes}\footnote{https://en.wikipedia.org/wiki/List\_of\_Malaysian\_dishes}\footnote{https://en.wikipedia.org/wiki/List\_of\_Thai\_dishes}\footnote{https://en.wikipedia.org/wiki/List\_of\_Vietnamese\_dishes}\footnote{https://en.wikipedia.org/wiki/List\_of\_Indian\_dishes}. Given a title, a rank list of 1 to 6,469 image-recipe pairs are retrieved. For example, the recipes ``nasi lemak ipin upin'', ``nasi lemak hijau pandan'' and ``sambal nasi lemak'' are retrieved by using ``nasi lemak" as the search keywords. In total, a dataset composed of 104,842 pairs was curated using 776 dish titles from five different cultures. Please refer to Section F for statistics on the crawled image-recipe pairs, as well as ingredient and action overlaps across different cultures.

To ensure data quality, we perform deduplication by removing samples with duplicate recipe titles from the test set. Specifically, we randomly retain one sample for each group of duplicate recipes and discard the rest. To account for the randomness in this process, we conduct 10 independent samplings and report the average performance. The statistics of training, validation, and testing sets are listed in Table~\ref{tab:cookpad_statistics}. The dataset is shared publicly\footnote{https://github.com/GZWQ/multilingual-image-recipe-retrieval}.

\subsection{Zero-shot Retrieval}

\input{tables/cookpad_comparison}

We conduct a zero-shot retrieval experiment to assess the robustness of the proposed method. To reduce the influence of visual-language models (e.g., multilingual CLIP variants) on the results, we use less popular dishes as query images.  We follow two protocols to ensure that the recipes in the testing set are both less popular and unseen in the training and validation sets. First, the testing set is composed of recipes that are retrieved with the search keywords different from the other two sets. Second, these keywords correspond to the dishes that are less popularly consumed. We verify the dish popularity in two steps: (1) prompting GPT-4o to rank the search keywords based on dish popularity in a particular culture, (2) sorting the keywords based on the number of returned recipes from Cookpad. Finally, the image-recipe pairs that are retrieved by the search keywords corresponding to the less popular dishes identified by both steps are included in the test set. All the images in the test set are used as search queries in the zero-shot experiment.

During retrieval, the culture of a query image needs to be known as a priori to activate the appropriate culture-specific module for representation debiasing. To this end, we train a classifier using the training data to predict the cultures of search queries. Table~\ref{tab:cookpad_comparison} lists the average retrieval performance for 18,256 search queries. Note that we experiment with three different backbones that support the native languages of five cultures for cross-modal retrieval: NLLB-SigLIP~\cite{visheratin2023nllb}, multilingual CLIP (mClip)~\cite{carlsson2022cross}, and OpenCLIP~\cite{ilharco_gabriel_2021_5143773}. Table~\ref{tab:cookpad_comparison} lists the retrieval performances of the backbones with different plugged-in debiasing models. For reference, we also list the oracle result, assuming the culture of a search query is known.

As shown in Table~\ref{tab:cookpad_comparison}, either the ingredient or action debiasing module contributes to performance improvement consistently across three multilingual CLIP variants. Combining both modules leads to the largest margin of improvement, for example,  elevating medR by about 15 and 3 ranks on the M-CLIP and OpenCLIP backbones, respectively. Although action debiasing on NLLB-SigLP with a classifier underperforms compared to the baseline, our proposed debiasing methods consistently improve both R@1 and R@10 on M-CLIP and OpenCLIP. Compared to the result for monolingual recipe retrieval on 10K and 20K test sets, the margin of improvement is larger. This basically indicates the benefit of mitigating representation bias for a dataset composed of multiple cuisines. Note that our result (``Classifier") is close to the oracle performance, even though the cultural predictions of the query images are suboptimal. For details on the performance of the culture prediction classifier, please refer to Section H in the supplementary. It is also worth mentioning that both debiasing modules introduce minimal overhead to the backbone model. For example, when using OpenClip with both debiasing modules, the retrieval speed for a single query is only 12 milliseconds. Please refer to Section I in the supplementary for a detailed training and inference times comparison across different models.

Table~\ref{tab:cookpad_comparison_country} further details the retrieval performances on different cultures. The baseline results (without debiasing) for Vietnamese and Indian cultures are relatively poor compared to other cultures. The effect of debiasing representation for these cultures is particularly effective, for example, by elevating the medR for about 100 ranks on the Indian culture when using mCLIP as the backbone. By debiasing the biases in ingredients and actions, Vietnamese and Indian cultures yield a relatively large margin of improvement. The MedR performances are somewhat correlated with the training data size. The training sets for Indian and Vietnamese cultures are smaller than Indonesia and Thailand, which result in higher values of medR. Although the training size for Malaysian culture is not larger than Vietnam, it benefits from the Indonesian training data for sharing similar dishes. Our results generally indicate that debiasing representation using our approach benefits low-resource cultures (e.g., Vietnam, India) more than mid or high-resource cultures (e.g., Thailand, Indonesia). Please see Section G in the supplementary for the full set of results, including recipe-to-image retrieval.

\input{tables/cookpad_qualitative}

Figure~\ref{fig:cookpad_ing_as_key} shows an example illustrating the benefit of debiasing representation related to both ingredients and actions. Given a query image of Indonesian culture, debiasing by either ingredient or action modules will result in an Indian culture recipe being returned as the top-1 result. By enhancing the query image representation with the ingredients (e.g., banana leaf) and cooking actions (e.g., boil and steam), the groundtruth recipe is retrieved. More examples can be found in Section J of the supplementary, including failure cases where dishes are covered by soup or obscured by toppings.

%% file: tables/cookpad_statistics.tex
\begin{table}[]
    \centering
    \caption{Multi-cultural cuisine recipe dataset.}
    \begin{adjustbox}{width=0.45\linewidth}
     \begin{tblr}{cccc}
\hline
         & Train & Val & Test \\ \hline
Indonesia      &  18,001    &  3,177   & 3,588   \\ \hline
Malaysia    &   13,099   &  2,312   &  3,437  \\ \hline
Thailand   &   16,833   &  2,971   &  3,977  \\ \hline
Vietnam  &  15,045    & 2,656    & 3,145   \\ \hline
India       &   10,618   &  1,874   & 4,109   \\ \hline
\textbf{Total} & \textbf{73,596} & \textbf{12,990} & \textbf{18,256}   \\ \hline 

\end{tblr}
\end{adjustbox}
    \label{tab:cookpad_statistics}
\end{table}

%% file: tables/cookpad_comparison.tex
\begin{table}
\centering
\caption{Performance of multicultural recipe retrieval. ``Oracle" assumes the culture of a search query is known. ``Classifier" predicts the culture of a query for retrieval.}
\resizebox{0.8\columnwidth}{!}{
\begin{tabular}{@{}l|cccc|cccc@{}}
\toprule
\multirow{2}{*}{} & \multicolumn{4}{c|}{\textbf{Oracle}} & \multicolumn{4}{c}{\textbf{Classifier}} \\ \cmidrule(l){2-9} 
                  & medR     & R@1      & R@5    & R@10    & medR     & R@1      & R@5    & R@10   \\ \midrule

NLLB-SigLIP~\cite{visheratin2023nllb}  &  176.9  &  5.2 &  12.9   & 17.9 &  176.9  &  5.2 &  12.9   & 17.9  \\  
\textbf{+Ingredient} & 165.2  &  5.4 &  13.3 &  18.5   & 168.3 & 5.1 & 13.1 & 18.2  \\
\textbf{+Action}  &  175.3  & 5.5  &  13.1   & 18.0   & 178.1 & 5.0 & 12.8 & 17.5 \\
\textbf{+Both}  & 151.3   &  6.0 &   14.1  &  19.4   & 153.5 & 5.6 & 13.6 &  18.7 \\ \midrule

M-CLIP~\cite{carlsson2022cross}   &  72.7  &  9.4 &  20.0   &  26.2 &  72.7  &  9.4 &  20.0   &  26.2  \\
\textbf{+Ingredient}  &  58.9  &  9.7 &  21.3   & 28.3  & 59.0  & 9.4 & 20.8 & 27.5 \\  
\textbf{+Action}  &  57.0  &  9.8 &  21.1   & 28.2  & 57.0 & 9.5 & 20.5 & 27.4 \\  
\textbf{+Both}  &  55.8  &  9.8 &  21.4   & 28.6 & 56.0 & 9.6 & 21.0 & 28.5 \\ \midrule

OpenCLIP~\cite{ilharco_gabriel_2021_5143773}  & 18.9 & 16.9 & 33.3 & 42.0 & 18.9 & 16.9 & 33.3 & 42.0  \\
\textbf{+Ingredient}  & 16.0 & 18.0 & 35.5 & 44.2 & 16.0 & 17.5 & 35.0 & 43.8\\ 
\textbf{+Action}   & 16.0 & 18.2 & 35.5 & 44.2 & 16.3 & 17.6 & 35.0 & 43.6 \\
\textbf{+Both}  & 15.4 & 18.4 & 35.8 & 44.5 & 16.0 & 18.0 & 35.1 & 44.1 \\ 
\bottomrule
\end{tabular}
}
\label{tab:cookpad_comparison}
\end{table}

\begin{table}
\caption{MedR performance for five cultures (ID: Indonesia, MY: Malaysia, TH: Thailand, VN: Vietnam, IN: India).}
\centering
\resizebox{0.65\columnwidth}{!}{
\begin{tabular}{@{}l|c|c|c|c|c@{}}
\toprule
& \textbf{ID} & \textbf{MY} & \textbf{TH} & \textbf{VN} & \textbf{IN} \\
\midrule
NLLB-SigLIP~\cite{visheratin2023nllb}   & 119.9 & 95.3 & 106.7 & 420.5 & 312.8 \\  
\textbf{+Ingredient}  & 115.3 & 86.9 & 83.6 & 426.9 & 301.9 \\
\textbf{+Action}      & 133.5 & 91.4 & 104.1 & 411.4 & 301.1 \\
\textbf{+Both}        & 113.8 & 79.9 & 87.3 & 329.3 & 300.5 \\ \midrule

M-CLIP~\cite{carlsson2022cross}    & 35.5 & 29.6 & 30.9 & 123.7 & 373.5 \\
\textbf{+Ingredient}  & 29.2 & 23.3 & 27.7 & 108.3 & 300.7 \\  
\textbf{+Action}      & 28.6 & 21.9 & 26.4 & 102.2 & 302.6 \\  
\textbf{+Both}        & 28.0 & 23.2 & 24.5 & 99.6 & 276.1 \\ \midrule

OpenCLIP~\cite{ilharco_gabriel_2021_5143773}   & 14.9 & 11.0 & 5.9 & 25.0 & 74.8 \\
\textbf{+Ingredient}  & 14.4 & 9.0 & 5.0 & 20.5 & 64.5 \\ 
\textbf{+Action}      & 14.4 & 9.7 & 5.0 & 20.1 & 60.9 \\
\textbf{+Both}        & 13.8 & 9.9 & 5.0 & 18.9 & 63.9 \\ 
\bottomrule
\end{tabular}
}
\label{tab:cookpad_comparison_country}
\end{table}

%% file: tables/cookpad_qualitative.tex
\begin{figure}
    \centering
    \includegraphics[width=0.7\linewidth]{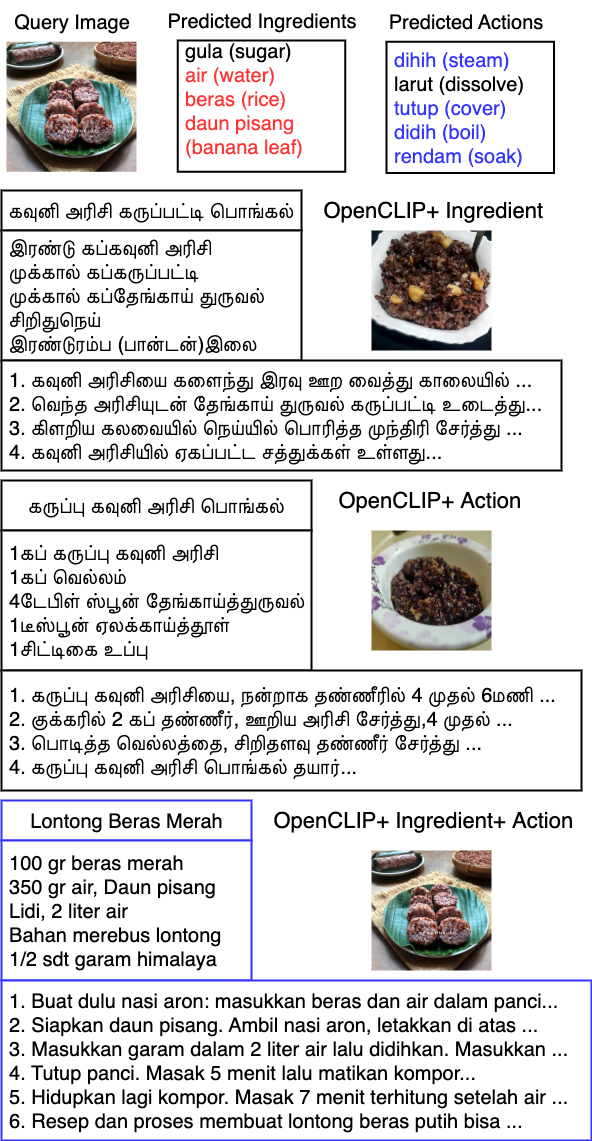}
    
    \caption{Example showing how the ingredient and action debiasing disambiguates similar recipes across cultures. The first row displays the query image, ingredients and actions predicted by debiasing modules. The following rows show the top-1 retrieved recipes (and the ground-truth images) by different debiasing modules. The ingredients and actions correctly predicted are marked in red and blue, respectively.}
    \label{fig:cookpad_ing_as_key}
\end{figure}

%% file: sections/7_conclusion.tex
\section{Conclusion}

Inspired by causal inference, we have presented a backdoor adjustment approach to alleviate the representation biases in cross-modal image-to-recipe training. Experimental results on both monolingual and multicultural datasets show noticeable retrieval improvement introduced by our proposed apprach. Particularly, debiasing biases due to both ingredients and actions lead to the largest margin of improvement. Furthermore, the results indicate that debiasing representation benefits retrieval more on the multicultural dataset than the monolingual dataset. The medR improvement is more pronounced for low-resource cultures (e.g., Vietnam, India) than for high-resource ones in our dataset.

%% file: sections/8_ack.tex
\section*{Acknowledgment}
This research / project is supported by the Ministry of Education, Singapore, under its Academic Research Fund Tier 2 (Proposal ID: T2EP20222-0046). Any opinions, findings and conclusions or recommendations expressed in this material are those of the author(s) and do not reflect the views of the Ministry of Education, Singapore.

%% file: sections/X_suppl.tex
% \clearpage
\nobalance
\appendix

\section*{Appendix}

\input{tables_sup/1_sota_comparison}

In this supplementary document, we first present additional results and analyses for the monolingual recipe retrieval task. This includes an extensive set of performance comparisons for both image-to-recipe and recipe-to-image retrieval tasks (Section~\ref{sec:sup_r2i}). We also provide detailed results from scalability tests (Section~\ref{sec:sup_scalability}) and ablation studies that explore the selection of ingredients and cooking actions for dictionary construction (Section~\ref{sec:sup_dictionary}). Furthermore, we include additional examples for qualitative and error analyses (Section~\ref{sec:sup_qualitative}). Finally, we provide the derivation of the debiasing equations (Section~\ref{sec:equation_derivation}).

For the multicultural recipe retrieval task, we begin by presenting additional statistics on the curated multicultural dataset (Section~\ref{sec:cookpad_dataset}), including data distribution and the overlap of ingredients and actions across cultures. This is followed by a comprehensive set of retrieval results for both image-to-recipe and recipe-to-image tasks (Section~\ref{sec:multilingual_r2i}). We also include the confusion matrix for the culture prediction classifier (Section~\ref{sec:confusion_matrix}), along with a comparison of model training and inference times (Section~\ref{sec:time_comparison}). Lastly, we present further examples for qualitative and error analyses (Section~\ref{sec:multi_culture_qualitative}).

\section{Recipe-to-Image Retrieval Results}\label{sec:sup_r2i}

Table~\ref{tab:comparison_full} presents the complete results for image-to-recipe and recipe-to-image retrieval after debiasing retrieval models using various confounders. For image-to-recipe retrieval, consistent improvements are observed across all models with the proposed debiasing methods. Ingredient-only debiasing achieves slightly greater gains than action-only debiasing, while debiasing both ingredients and actions yields the most significant improvements. In the recipe-to-image retrieval task, models like TFood and VLPCook show competitive performance with action-only debiasing compared to ingredient-only approaches. However, combining both ingredients and actions for bias removal during representation learning leads to further performance gains, with R1 improvements ranging from 0.9\% to 1.8\% compared to single-factor debiasing across the three baseline methods.

\section{Scalability Test} \label{sec:sup_scalability}
\input{tables_sup/2_sota_scalability}

We present the full results for both image-to-recipe and recipe-to-image retrieval tasks on larger test sets, ranging from 20K to 50K samples. Results for image-to-recipe are shown in Table~\ref{tab:scalability_full_i2r}, and for recipe-to-image in Table~\ref{tab:scalability_full_r2i}. For both retrieval tasks, the proposed debiasing module consistently enhances the performance of H-T~\cite{salvador2021revamping}, TFood~\cite{shukor2022transformer}, and VLPCook~\cite{shukor2022structured}.

\section{Ingredient and Cooking Action Dictionaries} \label{sec:sup_dictionary}
\input{tables_sup/3_dictionary}

Theoretically, all ingredients and cooking actions should be included during representation learning to eliminate bias. However, increasing the dictionary size complicates the training of ingredient and action generators, negatively affecting generation performance. This highlights a trade-off between retrieval and generation. To address this, we can optimize dictionary size by selecting a subset of ingredients and cooking actions that maximize retrieval performance while preserving generation accuracy.

We first investigate the impact of ingredient dictionary size on retrieval performance by debiasing H-T~\cite{salvador2021revamping} using ingredients only. The dictionary consists of popular ingredients from Recipe1M, including those that can become "invisible" during cooking (e.g., salt, butter). Intuitively, invisible ingredients are unlikely to be predicted from a food image and may be redundant in the dictionary. However, Table~\ref{tab:ingredient_dic_size} provides empirical insights into this intuition. First, we remove 250 invisible ingredients from the default dictionary of 500 ingredients, resulting in a slight impact on retrieval performance. Adding 250 more visible ingredients (based on frequency) to this reduced dictionary slightly improves retrieval performance but does not surpass the default dictionary containing both visible and invisible ingredients. This suggests that invisible ingredients still provide supplementary value in debiasing image representations. We attribute this to the ingredient classifier's ability to infer hard-to-see or invisible ingredients based on co-occurrence relationships in cooking~\cite{chen2020zero}. As shown in Table~\ref{tab:ingredient_dic_size}, smaller dictionaries generally reduce retrieval performance despite improving classification accuracy. Conversely, increasing the size to include the 1,000 most popular ingredients negatively impacts both classification and retrieval performance. A dictionary size of 500 ingredients strikes an effective balance in our experiments.

We fix the ingredient dictionary size at 500 and investigate the impact of action dictionary size on retrieval performance by debiasing H-T with both ingredients and cooking actions. Table~\ref{tab:action_dic_size} illustrates the impact of action dictionary size on the performance of both action generation and recipe retrieval. In this experiment, actions are sorted by their frequency in the Recipe1M training dataset, and only the most frequent actions are retained in the dictionary. As shown in Table~\ref{tab:action_dic_size}, a dictionary of 100 actions yields high classification accuracy but relatively low retrieval performance. In contrast, expanding the dictionary to 1,000 actions improves debiasing effects with a 1\% increase in recall@1 but results in a 2.3\% drop of F1 score in action generation. A smaller dictionary of 500 actions strikes a better balance, slightly outperforming the 1,000-action dictionary while requiring less memory, making it the optimal trade-off in our experiment.

\section{Qualitative Analysis Monolingual Recipe Retrieval} \label{sec:sup_qualitative}
\input{tables_sup/4_action_qualitative}
Debiasing both ingredients and cooking actions outperforms action-only debiasing in recipe retrieval, as it better distinguishes recipes that share similar sets of actions. Figure~\ref{fig:ret2action_analysis_2} illustrates an example where H-T+Action fails to differentiate recipes with similar cooking actions (e.g., preheat, spread, sprinkle, and bake). By incorporating predicted ingredients (e.g., red onions, parmesan cheese, tomato paste, and garlic cloves) alongside cooking actions, the ground-truth recipe is ranked in the Top-1 position.

% \section{Error Analysis}\label{sec:action_error_analysis}
\input{tables_sup/5_action_error_analysis}
When the transformative actions are misidentified, the action debiasing module may inadvertently harm retrieval performance. Figure~\ref{fig:action_error} illustrates examples where action debiasing leads to incorrect retrieval results. In the first example, although a sequence of preservative actions is correctly recognized, the transformative action of the query image is mistakenly identified as frying. This incorrect transformative action information, when incorporated into the image representation, causes the rank of the ground-truth recipe to drop from $3^{rd}$ (using H-T+ingredient) to $5^{th}$ (using H-T+ingredient+action). Similarly, in the second example, the actual transformative action is baking, but the prediction includes both baking and frying, which pulls baked-then-fried chicken dishes closer and pushes the ground-truth recipe's rank from $4^{th}$ to $11^{th}$.

\section{Debiasing Equation Derivation} \label{sec:equation_derivation}
In Section 3 of the main paper, we derive the similarity computation for recipe retrieval by approximating the backdoor adjustment. Here, we provide the complete details of the equation derivations for the approximation.
{\allowdisplaybreaks \begin{subequations} \label{eq:implement_equation1}
    \begin{align}
    & P(S|do(I),R) \notag \\
    & = \sum_{ing}\sum_{act} f_s(e_I,e_R, e_{ing}, e_{act})P(act|R, ing)P(ing|R) \label{eq:implement_eq1_act_supp} \\
    & = \mathbb{E}_{[ing|R]} \left[\mathbb{E}_{[act|R, ing]} \left[f_s(e_I,e_R, e_{ing}, e_{act})\right]\right] \label{eq:implement_eq2_act_supp} \\
    & = \mathbb{E}_{[ing|R]} \left[\mathbb{E}_{[act|R, ing]} \left[e_R \cdot (e_I + e_{ing} + e_{act})\right] \right] \label{eq:substitute_equation_act_supp} \\
    & = \mathbb{E}_{[ing|R]} \left[e_R \cdot (e_I + e_{ing} + \mathbb{E}_{[act|R, ing]}[e_{act}]) \right] \label{eq:implement_eq3_act_supp}  \\
    & = e_R \cdot \left(e_I + \mathbb{E}_{[ing|R]}[e_{ing}] + \mathbb{E}_{[ing|R]}[\mathbb{E}_{[act|R, ing]}[e_{act}] ]\right) \label{eq:implement_eq3_1_act_supp} \\
    &= e_R \cdot \Bigg(e_I + \sum_{ing} P(ing|R) \cdot e_{ing} \notag \\
&\quad + \sum_{ing} P(ing|R) \sum_{act} P(act|R, ing) \cdot e_{act} \Bigg) \label{eq:eq3_1_act_expand_expectation_supp} \\
& \approx e_R \cdot \Bigg(e_I + \sum_{ing} P(ing|I) \cdot e_{ing} \notag \\
&\quad + \sum_{ing} P(ing|I) \sum_{act} P(act|I, ing) \cdot e_{act} \Bigg) \label{eq:eq3_1_act_expand_expectation_approx_supp} \\
&= e_R \cdot \Bigg(e_I + \sum_{ing}P(ing|I) \cdot e_{ing} \notag \\
    &\quad + P(ing_1|I) \cdot \sum_{act}P(act|I, ing_1)\cdot e_{act} \notag \\
    &\quad + P(ing_2|I) \cdot \sum_{act}P(act|I, ing_2)\cdot e_{act}   \notag \\
    &\quad + \ldots + P(ing_K|I) \cdot \sum_{act}P(act|I, ing_K)\cdot e_{act} \Bigg) \label{eq:eq3_1_act_expand_expectation1_supp} 
    \end{align}
\end{subequations}}
where Eq.~(\ref{eq:implement_eq2_act_supp}) is derived according to the definition of expectation. As the similarity function $f_s$ is often implemented as a dot product operation, we set $f_s(e_I,e_R, e_{ing}, e_{act})=e_R \cdot (e_I + e_{ing} + e_{act})$ in Eq.~(\ref{eq:substitute_equation_act_supp}), where a similar implementation is also used by~\cite{wang2020visual,qi2020two}. Eq.~(\ref{eq:implement_eq3_act_supp}) and Eq.~(\ref{eq:implement_eq3_1_act_supp}) are obtained by moving the expectations inside the parentheses. Eq.~(\ref{eq:eq3_1_act_expand_expectation_supp}) is obtained based on the definition of expectation. Since $R$ is our search target during retrieval, we approximate $R$ with $I$ Eq.~(\ref{eq:eq3_1_act_expand_expectation_supp}), i.e., approximating $P(ing|R)$ and $ P(ing|R, ing)$ in Eq.~(\ref{eq:eq3_1_act_expand_expectation_supp}) with $P(ing|I)$ and $ P(ing|I, ing)$, respectively, and Eq.~(\ref{eq:eq3_1_act_expand_expectation_approx_supp}) is derived. After expanding $\sum_{ing} P(ing|I)\sum_{act} P(act|I, ing)\cdot e_{act}$ in Eq.~(\ref{eq:eq3_1_act_expand_expectation_approx_supp}), we have Eq.~(\ref{eq:eq3_1_act_expand_expectation1_supp}).

\section{Multicultural Recipe Cookpad Dataset}\label{sec:cookpad_dataset}

Table~\ref{tab:data_distrinution} shows the distribution of the dataset across different cultures, which is notably imbalanced. Indonesia has the highest number of query keywords and consequently the most crawled image-recipe pairs, while India has the fewest keywords and the least number of pairs. This disparity is influenced by the number of dish titles provided by Wikipedia and the corresponding number of successfully crawled image-recipe pairs.

\begin{table}[H]
    \centering
    \caption{Statistics of the crawled Cookpad dataset.}
    \resizebox{0.42\textwidth}{!}{
    \begin{tabular}{l l r r}
        \hline
        Culture & Query keywords & Crawled pairs & Percentage (\%) \\
        \hline
        Indonesia & 310 & 24,766 & 24 \\
        Malaysia  & 112 & 18,848 & 18 \\
        Thailand  & 182 & 23,781 & 22 \\
        Vietnam   & 102 & 20,846 & 20 \\
        India     & 70  & 16,601 & 16 \\
        \hline
    \end{tabular}
    }
    \label{tab:data_distrinution}
\end{table}

\input{tables_sup/11_cookpad_dataset}

We also present the overlap percentages (\%) in ingredients and actions are shown in Table~\ref{tab:ing_overlap} and Table~\ref{tab:act_overlap}, respectively. As shown, Indonesia and Malaysia share a high degree of overlap in both ingredients and cooking actions, while India exhibits the least overlap with other cultures in both categories.

\input{tables_sup/7_cookpad_r2i}

\section{Multilingual Recipe-to-Image Retrieval Results}\label{sec:multilingual_r2i}

\input{tables_sup/8_cookpad_region_wise_comparison}

Table~\ref{tab:comparison_both_directions} presents the results for image-to-recipe (I2R) and recipe-to-image (R2I) retrieval. The performance trends are similar, where similar degrees of improvements are introduced by different debiasing modules for both I2R and R2I. Table~\ref{tab:supp_cookpad_comparison} further details the performance of I2R for five different cultures.

\section{Culture Prediction Classifier Confusion Matrix}\label{sec:confusion_matrix}

In Table 5 of the main paper, we list both results: Oracle (with prior knowledge of culture being assumed) and Classifier (a classifier for predicting culture). To better explain our results in Table 5, we show the confusion matrix of the classifier in Table~\ref{tab:9_confusion_matrix}. As seen, while classification is suboptimal for some cultures, the retrieval result of Classifier is still close to that of Oracle. 

\input{tables_sup/9_confusion_matrix}

\section{Training and Testing Time Comparison}\label{sec:time_comparison}

We show the training time (minutes per epoch) for both monolingual and multilingual models in Table~\ref{tab:train_time}. None means no debiasing; Single means debiasing with either ingredients or cooking actions; Both means debiasing with both ingredients and actions. For the multicultural dataset, the number of training epochs is 30 for the models with and without debiasing. For the monolingual dataset, the number of epochs is 100, as the backbones used are weaker than OpenCLIP.

\begin{table}[H]
    \centering
        \caption{Training time comparison per epoch.}

    \begin{tabular}{llll}
\hline
         & None & Single  & Both \\ \hline
H-T~\cite{salvador2021revamping}      &   14min  &  26min  &   30min  \\ \hline
TFood~\cite{shukor2022transformer}    &   21min  &  94min  &  123min  \\ \hline
VLPCook~\cite{shukor2022structured}  &  77min   &  118min  &  134min  \\ \hline
OpenCLIP~\cite{ilharco_gabriel_2021_5143773}  &  17min  & 22min   &  25min  \\ \hline
\end{tabular}
    \label{tab:train_time}
\end{table}

The average inference time (including retrieval time) is presented in Table~\ref{tab:test_time} (milliseconds per image query). The overhead is considered acceptable. Even for large models such as VLPCook and OpenCLIP, our model (Both) can process 65 and 77 queries per second, respectively.

\input{tables_sup/10_time_comparison}

\section{Qualitative Analysis Multicultural Recipe Retrieval}\label{sec:multi_culture_qualitative}

\input{tables_sup/14_cam_visulization}

We present an example in Figure~\ref{fig:cam_visulization} to illustrate how our debiasing modules help the model attend to different image regions, thereby improving retrieval performance. As shown, although ``soy sauce'' is barely visible in the dish image, its presence affects the color of the pork due to pickling, and the model correctly associates the prediction of ``soy sauce'' with the pork, as highlighted in the activation map. Similarly, the action of ``chopping'' is localized to the shallots, which have undergone this preparation. Without debiasing, the activation maps often fail to capture such fine-grained ingredient and action-level cues.

\input{tables_sup/13_cam_cookpad}
We present two examples of distinguishing visually similar recipes using our proposed debiasing method in Figure~\ref{fig:cam_cookpad}. In the first example, the two recipes share many common ingredients, such as chicken wings, garlic, and salt. Despite their visual similarity and overlapping ingredients, the query image corresponds to a dish of chicken sticky rice, which uses sticky rice as a key ingredient, whereas the visually similar dish is chicken rice, made with regular rice. Our debiasing model correctly predicts the presence of sticky rice, enabling it to distinguish between these two visually similar recipes. Additionally, we provide the class activation map for ``sticky rice'' in Figure~\ref{fig:cam_cookpad}(c), which highlights the rice regions in the image, indicating the model's attention to the relevant area.

In the second example, the debiasing module identifies ingredients unique to the query image such as lime juice and mint leaves, which helps distinguish it from a visually similar recipe that shares ingredients like steak, garlic, and peppers. Although lime juice is not directly visible in the image, the class activation map shows that its prediction is based on attention to the entire dish, which aligns with the fact that the juice is mixed throughout the food.

\input{tables_sup/12_cookpad_error}

Figure~\ref{fig:cookpad_error} presents examples where the debiasing module fails to distinguish between recipes with similar visual appearances. In the first case, ``Vegetable Cream Soup'' and ``KFC-style Cream Soup'' appear visually similar and share most ingredients. However, ``Vegetable Cream Soup'' uses ``cornstarch'', while ``KFC-style Cream Soup'' contains ``flour''. The model, after debiasing, incorrectly predicts ``flour'' for the image of ``Vegetable Cream Soup'', reducing the correct recipe’s rank from 4 to 10 and mistakenly selecting ``KFC-style Cream Soup'' as the top-1 result. In the second example, although most ingredients are correctly identified, the model incorrectly predicts the cooking action as baking for ``Fried Chicken with Coconut Serundeng''. This leads it to favor ``Grilled Chicken'', which involves baking, thereby pushing the correct recipe from rank 3 to 10.

%% file: tables_sup/1_sota_comparison.tex
\begin{table*}[ht]

\centering
\caption{Comparison on 1k and 10k test sets. medR ($\downarrow$), Recall@k ($\uparrow$) are reported. The proposed debiasing successfully boosts the performance of existing cross-modal retrieval methods (H-T, TFood, VLPCook), especially on the 10k set.}
% \vspace{0.2cm}
\resizebox{\textwidth}{!}{
\begin{tabular}{@{}l|cccc|cccc|cccc|cccc@{}}
\toprule
\multirow{3}{*}{} & \multicolumn{8}{c|}{\textbf{1k}}      & \multicolumn{8}{c}{\textbf{10k}} \\ \cmidrule(l){2-17} 
                  & \multicolumn{4}{c|}{\textbf{image-to-recipe}} & \multicolumn{4}{c|}{\textbf{recipe-to-image}} & \multicolumn{4}{c|}{\textbf{image-to-recipe}} & \multicolumn{4}{c}{\textbf{recipe-to-image}} \\ \cmidrule(l){2-17} 
                  & medR     & R@1      & R@5    & R@10    & medR     & R@1     & R@5     & R@10    & medR     & R@1      & R@5    & R@10    & medR     & R@1     & R@5     & R@10    \\ \midrule
% Salvador et al. \cite{salvador2017learning}    & 5.2      & 24.0    & 51.0      &   65.0     &   5.1       & 25.0       &     52.0   &       65.0 & 41.9    & -     &  -     &   -     &      39.2    &      -  &     -   &    -    \\

RIVAE~\cite{kim2021learning}  & 2.0      & 39.0    & 70.0      &   79.0     &   -       & -       &     -   &       - & -   & -     &  -     &   -     &    -   &      -  &     -   &    -    \\

% Adamine \cite{carvalho2018cross}   &    1.0      &   39.8     &   69.0     &   77.4  &    1.0      & 40.2    &  68.1     &    78.7     &      13.2    &   14.9     &    35.3    &    45.2 & 12.2     & 14.8    &   34.6    &  46.1   \\
R2GAN \cite{zhu2019r2gan}                  &     2.0     &     39.1    &   71.0    &  81.7      &  2.0        &    40.6    &    72.6    &       83.3 &  13.9        &   13.5      &   33.5    &     44.9   &     12.6     &   14.2     &   35.0     &   46.8     \\
MCEN \cite{fu2020mcen}               &   2.0       &   48.2      &    75.8   &      83.6  &    1.9      &   48.4     &  76.1      &    83.7    &  7.2        &      20.3   &    43.3   &   54.4     &     6.6     &   21.4     &  44.3      &    55.2    \\
ACME \cite{wang2019learning}              &    1.0      &   51.8      &   80.2    &    87.5    &   1.0       &    52.8    &   80.2     &       87.6 &     6.7     &   22.9     &  46.8     &    57.9    &      6.0    &  24.4      &    47.9    &      59.0  \\
SN \cite{zan2020sentence}               &    1.0      & 52.7        &     81.7  &    88.9    &    1.0      &  54.1      &  81.8      &   88.9     &    7.0      &    22.1    &    45.9   &    56.9    &    7.0      &     23.4   &  47.3      &  57.9      \\
IMHF \cite{li2021cross}               &    1.0      & 59.4        &     81.0  &    87.4    &    1.0      &  61.2      &  81.0      &   87.2     &    3.5      &    36.0    &    56.1   &    64.4    &    3.0      &     38.2   &  57.7      &  65.8      \\
% \textit{Wang et. al}  \cite{wang2021learning}               &    1.0      & 53.5        &     81.5  &    88.8    &    1.0      &  55.0      &  82.0      &   88.8     &    6.0      &    23.4    &    48.8   &    60.1    &    5.6      &     24.6   &  50.0      &  61.0      \\
SCAN \cite{wang2021cross}               &    1.0      & 54.0        &     81.7  &    88.8    &    1.0      &  54.9      &  81.9      &   89.0     &    5.9      &    23.7    &    49.3   &    60.6    &    5.1      &     25.3   &  50.6      &  61.6      \\
HF-ICMA \cite{li2021hybrid}               &    1.0      & 55.1        &     86.7  &    92.4    &    1.0      &  56.8      &  87.5      &   93.0     &    5.0      &    24.0    &    51.6   &    65.4    &    4.2      &     25.6   &  54.8      &  67.3      \\

MSJE \cite{xiez2021learning}               &    1.0      & 56.5        &     84.7  &    90.9    &    1.0      &  56.2      &  84.9      &   91.1     &    5.0      &    25.6    &    52.1   &    63.8    &    5.0      &     26.2   &  52.5      &  64.1      \\

SEJE \cite{xie2021learning}               &    1.0      & 58.1        &     85.8  &    92.2    &    1.0      &  58.5      &  86.2      &   92.3     &    4.2      &    26.9    &    54.0   &    65.6    &    4.0      &     27.2   &  54.4      &  66.1      \\

M-SIA \cite{li2021multi} &  1.0       &     59.3    &    86.3   &    92.6    &     1.0     &     59.8   &    86.7    &      92.8  &     4.0    &    29.2     &    55.0   &    66.2    &    4.0      &     30.3   &     55.6   &  66.5     \\

RDE-GAN \cite{sugiyama2021cross}               &    1.0      & 55.1        &     86.7  &    92.4    &    1.0      &  56.8      &  87.5      &   93.0     &    5.0      &    24.0    &    51.6   &    65.4    &    4.2      &     25.6   &  54.8      &  67.3      \\

X-MRS \cite{guerrero2021cross}             &  1.0       &     64.0    &    88.3   &    92.6    &     1.0     &     63.9   &    87.6    &      92.6  &     3.0    &    32.9     &    60.6   &    71.2    &    3.0      &     33.0   &     60.4   &  70.7       \\

Cooking Program ~\cite{papadopoulos2022learning}  &   1.0    &  66.8   &  89.8     &   94.6      &    -   &      -  &     -   &    - &    -   &      -  &     -   &    - &    -   &      -  &     -   &    -    \\

FARM~\cite{wahed2024fine}  &   1.0    &  73.7   &   90.7    &    93.4     &    1.0   &      73.6  &     90.8   &    93.5 &    2.0   &      44.9  &     71.8   &    80.0 &    2.0   &      44.3  &     71.5   &    80.0    \\
CREAMY~\cite{zou2024creamy}  &   1.0    &  73.3   &   92.5    &    95.6     &    1.0   &      73.2  &     92.5   &    95.8 &    2.0   &      44.6  &     71.6   &    80.4 &    2.0   &      45.0  &     71.4   &    80.0    \\

CIP~\cite{huang2023improving}  &    1.0     &   77.1   & 94.2   &   97.2     &  1.0    &   77.3     &    94.4   &     97.0   &  2.0     &   44.9    &   72.8  &  82.0    &  2.0   &   45.2   &    73.0    &   81.8   \\

DAR~\cite{song2024DAR}  &   1.0    &   77.3  &   95.3    &   97.7      &    1.0   &      77.1  &     95.4   &    97.9 &    2.0   &      47.8  &     75.9   &    84.3 &    2.0   &      47.4  &     75.5   &    84.1    \\

FMI~\cite{zhao2025cross}  &   1.0    &   77.4  &   95.8    &   97.6      &    1.0   &      77.1  &     95.4   &    97.7 &    1.0   &      48.4  &     76.3   &    81.9 &    1.0   &      49.5  &     79.2   &    83.1    \\\midrule

H-T~\cite{salvador2021revamping}            &  1.0  &  61.8 &   88.0  & 93.2 & 1.0 & 62.1   & 88.3  &  93.5     &    3.95  &  29.9 & 58.3  &  69.6 & 3.6 &   30.4    &  58.6     &      69.7     \\ 
\textbf{+Ingredient}           &  1.0  & 65.7  &  89.8   & 94.1 & 1.0 &  66.0  & 89.9  &  94.2     &   3.0   &  34.4 &  62.9 &  73.6 & 3.0 &  34.7     &   63.2    &   73.7              \\
\textbf{+Action}           &  1.0  &  63.6 &  88.1   & 92.6 & 1.0 &  63.3  & 88.5  &   92.9    &   3.0   & 32.1 &  60.3 & 71.1  & 3.0 &    32.4   &   60.1    &   70.9              \\ 
\textbf{+Both}           &  1.0  & 65.7  &  88.8   & 93.6 & 1.0 &  65.9  &  89.3 &   94.0    &   3.0   & 35.5 & 63.8  & 74.1  & 3.0 &   36.5    &    64.2   &   74.3              \\ \midrule

TFood~\cite{shukor2022transformer}           &  1.0  &  72.4 &   92.5  & 95.4 & 1.0 &  72.5  & 92.1  &   95.3    &   2.0   & 43.9  & 71.7  & 80.8  & 2.0 &    43.7   &    71.6   &      80.6           \\ 
\textbf{+Ingredient}            &  1.0  &  74.5   & 93.2 & 96.1 &  1.0  &  73.7 &  93.1     &  96.0    & 2.0  & 45.6  &  73.0 & 81.6 &   2.0    &   44.9   &   72.7  &  81.5              \\
\textbf{+Action}           &  1.0  & 73.8  &  93.1   & 95.8 & 1.0 &  73.6  &  93.1 &   96.0    &   2.0   & 45.1 &  72.6 &  81.3 & 2.0 &   45.6    &   72.8    &    81.3             \\ 
\textbf{+Both}           &  1.0  & 75.8  &  93.6   & 96.3 & 1.0 &  76.3  & 94.0  &   96.6    &   2.0   & 46.9 &  74.4 & 82.8  & 2.0 &  47.4     &  74.8     &   83.2              \\ \midrule

VLPCook~\cite{shukor2022structured}           &  1.0  &  77.4 &   94.8  & 97.1 & 1.0 &  78.0  & 94.9  &   97.1    &   2.0   & 48.8  & 76.2  & 84.5  & 1.6 & 49.9 &  76.9     &  85.0                      \\ 
\textbf{+Ingredient}            &  1.0  & 78.3  &  95.1   & 97.4 & 1.0 &  78.6  &  95.2 &  97.4    &  1.4    & 50.2  & 77.3 & 85.2  & 1.0 & 51.0 &  77.9   &  85.6                     \\ 
\textbf{+Action}           &  1.0  & 77.9  &  95.0   & 97.4 & 1.0 &  79.0  &  95.4 &   97.8    &   1.5   & 50.0 &  77.4 & 85.4  & 1.0 &   51.3    &    78.1   &  85.8               \\ 
\textbf{+Both}           &  1.0  &  79.1 &   94.6  & 97.0 & 1.0 &  78.3  & 95.0  &   97.2    &   1.0   & 51.7 & 78.2  & 85.9  & 1.0 &   52.2    &    78.4   &   86.0 \\             
% \\ \midrule

% Oracle           &  1.0  &  99.0 &  99.8   & 99.9 & 1.0 & 98.9   &  99.8 &  99.9     &  1.0    & 96.2 & 99.2  &  99.5 & 1.0 & 96.1 & 99.1      &     99.5                 \\ 
\bottomrule
\end{tabular}
}

\label{tab:comparison_full}
\end{table*}

%% file: tables_sup/2_sota_scalability.tex
\begin{table*}[th]
\centering
\caption{Scalability test on 20k, 30k, 40k and 50k test set for the image-to-recipe retrieval task.}
% \vspace{0.2cm}
\resizebox{\textwidth}{!}{
\begin{tabular}{@{}l|cccc|cccc|cccc|cccc@{}}
\toprule
\multirow{2}{*}{} %& \multicolumn{8}{c|}{\textbf{1k}}                                                    & \multicolumn{8}{c}{\textbf{10k}} \\ \cmidrule(l){2-17} 
                  & \multicolumn{4}{c|}{\textbf{20k}} & \multicolumn{4}{c|}{\textbf{30k}} & \multicolumn{4}{c|}{\textbf{40k}} & \multicolumn{4}{c}{\textbf{50k}} \\ \cmidrule(l){2-17} 
                  & medR     & R@1      & R@5    & R@10    & medR     & R@1     & R@5     & R@10    & medR     & R@1      & R@5    & R@10    & medR     & R@1     & R@5     & R@10    \\ \midrule

H-T~\cite{salvador2021revamping}            &  6.3  &  22.2 &  47.0   & 58.8 & 9.0 &  18.4  & 41.1  &  52.5     &  12.0    & 16.0 &  36.9 & 47.9  & 15.0 & 14.3 &  33.8     &  44.4  \\ 
\textbf{+Ingredient}          &  5.0  & 26.2  &  52.4   & 63.7 & 7.0 &  22.0  &  46.2 &  57.7     &   9.0   & 19.3 &  41.9 & 53.2  & 11.0 & 17.4 &   38.7    &    49.6       \\ 
\textbf{+Action}          &  5.8  &  24.3 &  49.7   & 60.9 & 8.0 &  20.3 &  43.7 &   54.8    &   10.0   & 17.8 & 39.5  & 50.4  & 12.6 & 15.9 & 36.3      &  47.1        \\ 
\textbf{+Both}          &  4.7  &  27.3 &  53.3   & 64.3 & 6.0 & 23.0  & 47.4  &   58.4    &   8.0   & 20.3 &  43.4 & 54.2  & 10.0 & 18.2 &  40.2     &  50.7        \\ \midrule

TFood~\cite{shukor2022transformer}           &  3.0  & 35.5  &  62.0   & 72.5 &  4.0 &  30.9  &  56.0 & 66.7      &   5.0   & 27.8 &  52.2 &  62.8 & 6.0  & 25.7 & 49.1  &  59.7       \\ 
\textbf{+Ingredient}            &  3.0  & 37.6  &  64.3   & 73.9 & 3.0 &  32.9  & 58.6  &  69.0     &   4.0   & 29.9 &  54.5 & 65.1  & 5.0  & 26.9 & 51.2  & 61.5         \\ 
\textbf{+Action}          &  3.0  & 36.3  &  63.5   & 73.4 & 4.0 & 31.6  & 57.8  &   68.1    &   4.0   & 28.5 &  53.7 &  64.3 & 5.0 & 26.2 &   50.5    &  61.2        \\ 
\textbf{+Both}          &  2.0  & 38.6  &  65.5   & 75.4 & 3.0 & 33.6  & 59.8  &   70.0    &   4.0   & 30.4 & 55.6  &  66.2 & 5.0 & 28.1 &    52.5   &     63.2     \\ \midrule

VLPCook~\cite{shukor2022structured}          &  2.0  &  40.2 &  67.4   & 77.2 & 3.0 &   35.2 &  61.6 &  72.2     &   4.0   & 32.0 &  57.5 &  68.4 &  4.0 & 29.7 & 54.5  & 65.3                  \\ 
\textbf{+Ingredient}          &  2.0  & 41.7  &  69.1   & 78.5  &  3.0  &  36.9 &   63.5    &   73.5   & 3.0 & 33.7  & 59.7 & 69.9 & 4.0  & 31.1 & 56.4  & 66.7       \\ 
\textbf{+Action}          &  2.0  &  41.0 &  68.4   & 78.1  & 3.0 & 36.0  &  62.7 &   73.0    &   3.0   & 32.7 & 58.6  & 69.1  & 4.0 & 30.2 &  55.5     &  66.1        \\ 
\textbf{+Both}          &  2.0  &  42.7 &  69.7 & 78.8 & 3.0 & 37.7  & 64.4  &  74.2     &   3.0   & 34.5 & 60.4  & 70.6  & 4.0 & 32.0 &   57.4    &  67.7        \\ 
% \midrule

% Oracle           &  1.0  & 94.6  &   98.8  & 99.2 & 1.0 & 93.5   & 98.4  &    99.0   &  1.0    & 92.6 & 98.0  & 98.8  & 1.0 & 91.8 &    97.7   &  98.7                    \\ 
\bottomrule
\end{tabular}
}

\label{tab:scalability_full_i2r}
\end{table*}

\begin{table*}[th]
\centering
\caption{Scalability test on 20k, 30k, 40k and 50k test set for the recipe-to-image retrieval task.}
% \vspace{0.2cm}
\resizebox{\textwidth}{!}{
\begin{tabular}{@{}l|cccc|cccc|cccc|cccc@{}}
\toprule
\multirow{2}{*}{} %& \multicolumn{8}{c|}{\textbf{1k}}                                                    & \multicolumn{8}{c}{\textbf{10k}} \\ \cmidrule(l){2-17} 
                  & \multicolumn{4}{c|}{\textbf{20k}} & \multicolumn{4}{c|}{\textbf{30k}} & \multicolumn{4}{c|}{\textbf{40k}} & \multicolumn{4}{c}{\textbf{50k}} \\ \cmidrule(l){2-17} 
                  & medR     & R@1      & R@5    & R@10    & medR     & R@1     & R@5     & R@10    & medR     & R@1      & R@5    & R@10    & medR     & R@1     & R@5     & R@10    \\ \midrule

H-T~\cite{salvador2021revamping}            &  6.0  &  22.9 &  47.8   & 59.3 & 9.0 & 19.1   & 41.8  &  53.0   &  11.2    &  16.6 &  37.5 & 48.5 & 14.0 & 14.8 &  34.3 &  45.1 \\ 
\textbf{+Ingredient}          &  5.0  & 26.7  &  52.6   & 63.9 & 7.0 &  22.5  & 46.6  &  58.0  &   8.8   & 19.8 &  42.3  & 53.5 & 10.0 & 17.9 &  39.1 & 50.1   \\ 
\textbf{+Action}          &  5.9  & 24.6  &   49.6 & 60.9 &  8.0    &  20.7 & 43.6 &  54.9  &  10.0  &  18.1 &  39.5  & 50.4 &  12.9    & 16.3 & 36.4 & 47.0   \\ 
\textbf{+Both}           &  4.0  &  28.4 &  53.9   & 64.7 & 6.0   & 24.2  & 48.0  & 58.7  &  8.0 & 21.6  &  44.1   & 54.7 &  10.0  & 19.6  & 40.9  &  51.4  \\  \midrule

TFood~\cite{shukor2022transformer}    &  3.0  & 35.6  &  62.2   & 72.5 & 4.0 &  31.0  & 56.6  &  67.2   &   5.0   & 28.0 &  52.3 &  63.0 & 6.0 & 25.8 &  49.1     & 59.8  \\ 
\textbf{+Ingredient}             &  3.0  & 37.0  &  63.9   & 73.8 & 3.0 &  32.4  & 58.3  &  68.7 &   4.0   & 29.3 &  54.4 & 64.9 & 5.0 & 27.0 &   51.2    &    61.7     \\ 
\textbf{+Action}          &  3.0  & 37.2  &  64.8  & 73.4 &  3.1    & 32.4  & 58.2 & 68.4  &  4.0  & 29.2  &  54.1  & 64.4 &   5.0   & 26.9  & 51.0 & 61.5 \\ 
\textbf{+Both}          &   2.1 &  38.8 &  65.6   & 75.4 &  3.0  &  34.1 & 60.2  &  70.4    &  4.0  & 30.8  &  56.1   & 66.5 &  5.0  &  28.5 &  53.0 &  63.5 \\ \midrule

VLPCook~\cite{shukor2022structured}   &  2.0  &  41.4 &  68.6   & 78.2 & 3.0 & 36.3   & 62.8  &   73.0 &   3.0   & 33.1 &  58.8 &  69.3 & 4.0 & 30.6 &  55.6     &  66.2   \\ 
\textbf{+Ingredient}       &  2.0  & 42.5  &  69.3   & 78.6 & 3.0 & 37.6   &  64.1 &  73.9  & 3.0 & 34.3  & 60.0 & 70.2 & 4.0  & 31.9 & 57.0  & 67.3    \\ 
\textbf{+Action}          & 2.0   &  42.4 &  69.6  &  79.0 &   3.0   & 37.5  & 64.1 &  74.0 &  3.0  & 34.2  &  59.9  & 70.3 &  4.0    & 31.8  & 56.7 &  67.4  \\ 
\textbf{+Both}        &  2.0  &  43.3 &  70.3   & 79.4 & 2.4   &  38.4 & 64.8  &  74.5 &  3.0  & 35.2  & 60.8 & 70.9 &  4.0  & 32.8  &  57.7 & 68.0 \\ 

% \midrule

% Oracle         &  1.0  & 94.5  &  98.7   & 99.2 & 1.0 &  93.2  &  98.3 &  99.0  &  1.0    & 92.2 &  98.0 &  98.8 & 1.0 & 91.3 &  97.7     &  98.6     \\ 
\bottomrule
\end{tabular}
}
\label{tab:scalability_full_r2i}
\end{table*}

%% file: tables_sup/3_dictionary.tex
\begin{table}[]
    \centering
    \caption{Impact of dictionary size and visibility of ingredients (on size of 500 ingredients). The table shows the ingredient classification and retrieval performances for H-T+Ingredient on 10k test size. Note that the columns marked with (visible only) show the results of using a dictionary that includes only ingredients that will likely be visible in a final cooked dish.}
    \begin{adjustbox}{width=0.85\columnwidth}

\begin{tabular}{ccccc}
\hline
\multirow{2}{*}{Size} & \multicolumn{3}{c}{Classification} & \multirow{2}{*}{Recall@1} \\ \cline{2-4}
                      & Precision    & Recall    & F1      &                           \\ \hline
100                   & 35.6         & 49.0      & 41.2    & 32.2                      \\ \hline
250 (Visible only) & 30.8 & 37.5 & 33.8 &  34.0 \\ \hline 

500                   & 30.7         & 38.1      & 34.0    & 34.4                      \\ \hline
500 (Visible only) & 29.1 & 33.9 & 31.3 & 34.3  \\ \hline 

1000                  & 29.7         & 35.2      & 32.2    & 34.0                      \\ \hline
\end{tabular}
   
    \end{adjustbox}{}
     
        \label{tab:ingredient_dic_size}
\end{table}

\begin{table}[]
    \centering
    
     \caption{Impacts of dictionary size on action classification and recipe retrieval for H-T+Both on 10k test size.}
\begin{adjustbox}{width=0.71\columnwidth}
\begin{tabular}{ccccc}
\hline

\multirow{2}{*}{Size} & \multicolumn{3}{c}{Classification} & \multirow{2}{*}{Recall@1} \\ \cline{2-4}
                      & Precision    & Recall    & F1      &                           \\ \hline
100                   &   0.482       &   0.325    &  0.388   &   34.3                    \\ \hline
500                   &    0.471      &   0.303    &   0.368  &    35.5                   \\ \hline
1000                  &     0.464     &  0.300     &  0.365   &   35.3                   \\ \hline
\end{tabular}
\end{adjustbox}
        \label{tab:action_dic_size}
\end{table}

%% file: tables_sup/4_action_qualitative.tex
\begin{figure*}
    \centering
    \includegraphics[width=\linewidth]{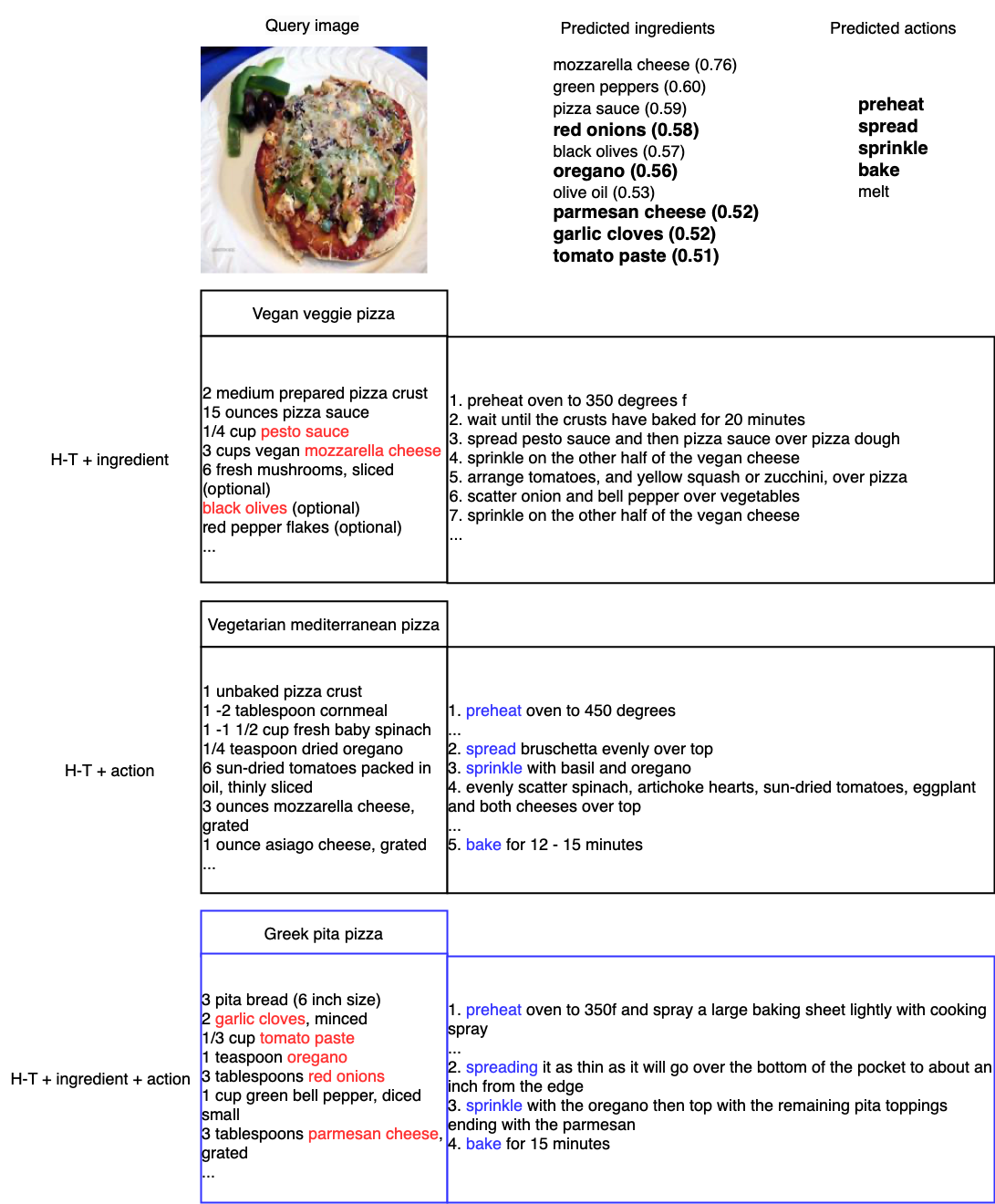}
    
    \caption{An example showing how the ingredient debiasing module disambiguates recipes with a similar set of actions. The first row displays the query image, predicted ingredients, and predicted actions. The following rows are the retrieved recipes by H-T+Ingredient, H-T+Action, and H-T+both, respectively. The correctly predicted ingredients and cooking actions are bolded. The predicted ingredients and cooking actions are marked in red and blue, respectively, in the recipes. The ground-truth recipe is boxed in blue.}
    \label{fig:ret2action_analysis_2}
\end{figure*}

%% file: tables_sup/5_action_error_analysis.tex
\begin{figure*}
    \centering
    \includegraphics[width=\linewidth]{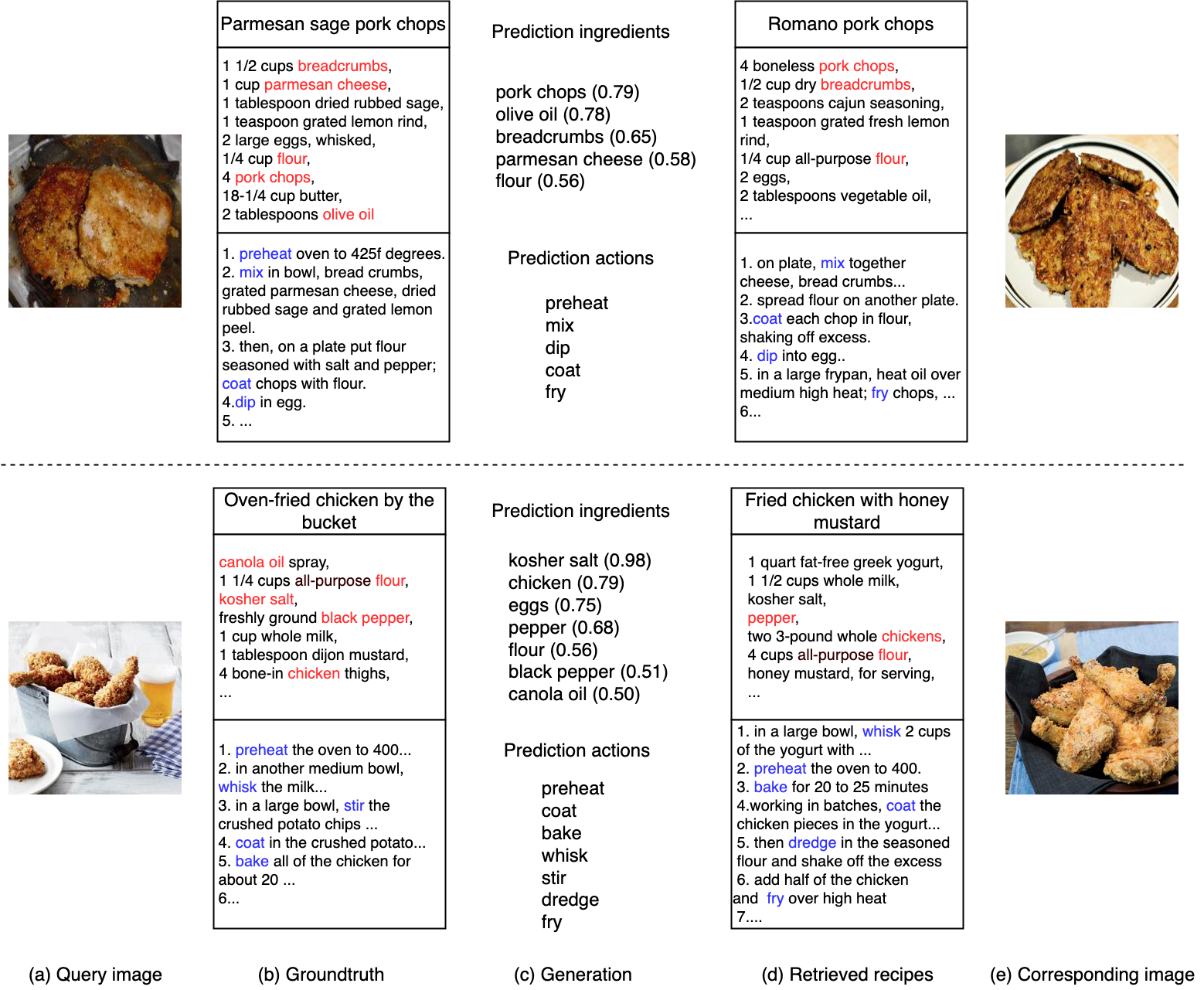}
    \caption{Failure examples of using action debiasing: query image (a), the corresponding ground-truth recipe (b), predicted ingredients and actions (c), retrieved recipe and its associated image by debiasing H-T with both ingredient and action (d) and (e).}
    \label{fig:action_error}
\end{figure*}

%% file: tables_sup/11_cookpad_dataset.tex
\begin{table}[H]
    \centering
    \caption{Pairwise ingredient overlap percentages (\%) between cultures.}
    \resizebox{0.4\textwidth}{!}{
    \begin{tabular}{l r r r r r}
        \hline
        & Indonesia & Malaysia & Thailand & Vietnam & India \\
        \hline
        Indonesia & 100 & 39 & 31 & 36 & 21 \\
        Malaysia  & 39  & 100 & 26 & 29 & 23 \\
        Thailand  & 31  & 26  & 100 & 31 & 13 \\
        Vietnam   & 36  & 29  & 31  & 100 & 18 \\
        India     & 21  & 23  & 13  & 18  & 100 \\
        \hline
    \end{tabular}
    }
    \label{tab:ing_overlap}
\end{table}

\begin{table}[H]
    \centering
    \caption{Pairwise action overlap percentages (\%) between cultures.}
    \resizebox{0.4\textwidth}{!}{
    \begin{tabular}{l r r r r r}
        \hline
        & Indonesia & Malaysia & Thailand & Vietnam & India \\
        \hline
        Indonesia & 100 & 47 & 35 & 34 & 31 \\
        Malaysia  & 47  & 100 & 43 & 31 & 31 \\
        Thailand  & 35  & 43  & 100 & 35 & 32 \\
        Vietnam   & 34  & 31  & 35  & 100 & 29 \\
        India     & 31  & 32  & 32  & 29  & 100 \\
        \hline
    \end{tabular}
    }
    \label{tab:act_overlap}
\end{table}

%% file: tables_sup/7_cookpad_r2i.tex
\begin{table*}
\centering
\caption{Comparison on the full test sets (size = 18,256) for both image-to-recipe and recipe-to-image retrieval tasks. medR ($\downarrow$), Recall@k ($\uparrow$) are reported. The ``Oracle'' setting assumes known cultural origin per image, enabling culture-specific debiasing. The ``Classifier'' setting predicts the culture origin first and then performs the corresponding culture debiasing.}
\resizebox{\textwidth}{!}{
\begin{tabular}{@{}l|cccc|cccc||cccc|cccc@{}}
\toprule
& \multicolumn{8}{c||}{\textbf{Image-to-Recipe}} & \multicolumn{8}{c}{\textbf{Recipe-to-Image}} \\ \cmidrule(lr){2-9} \cmidrule(l){10-17}
& \multicolumn{4}{c|}{\textbf{Oracle}} & \multicolumn{4}{c||}{\textbf{Classifier}} 
& \multicolumn{4}{c|}{\textbf{Oracle}} & \multicolumn{4}{c}{\textbf{Classifier}} \\
\cmidrule(lr){2-5} \cmidrule(lr){6-9} \cmidrule(lr){10-13} \cmidrule(l){14-17}
& medR & R@1 & R@5 & R@10 & medR & R@1 & R@5 & R@10 
& medR & R@1 & R@5 & R@10 & medR & R@1 & R@5 & R@10 \\
\midrule

NLLB-SigLIP~\cite{visheratin2023nllb}  
& 176.9 & 5.2 & 12.9 & 17.9 & 176.9 & 5.2 & 12.9 & 17.9   
& 156.5 & 5.7 & 13.6 & 19.1 & 156.5 & 5.7 & 13.6 & 19.1  \\

\textbf{+Ingredient} 
& 165.2 & 5.4 & 13.3 & 18.5 & 168.3 & 5.1 & 13.1 & 18.2 
& 148.2 & 5.9 & 14.5 & 19.9 & 153.1 & 5.7 & 14.2 & 19.6 \\

\textbf{+Action}  
& 175.3 & 5.5 & 13.1 & 18.0 & 178.1 & 5.0 &12.8  &17.5  
& 163.0 & 6.0 & 14.3 & 19.3 & 168.3 & 5.6 & 14.0 & 19.0 \\

\textbf{+Both}  
& 151.3 & 6.0 & 14.1 & 19.4 & 153.5 & 5.6 & 13.6 & 18.7  
& 135.6 & 6.6 & 15.3 & 20.7 & 139.0 & 6.2 & 14.5 & 19.9 \\

\midrule

M-CLIP~\cite{carlsson2022cross}  
& 72.7 & 9.4 & 20.0 & 26.2 & 72.7 & 9.4 & 20.0 & 26.2  
& 73.7 & 8.3 & 19.0 & 25.2 & 73.7 & 8.3 & 19.0 & 25.2 \\

\textbf{+Ingredient}  
& 58.9 & 9.7 & 21.3 & 28.3 & 59.0 & 9.4 & 20.8 & 27.5  
& 60.9 & 8.9 & 20.1 & 27.0 & 61.3 & 8.6 & 19.5 & 26.5 \\

\textbf{+Action}  
& 57.0 & 9.8 & 21.1 & 28.2 & 57.0 & 9.5 & 20.5 & 27.4 
& 60.7 & 8.9 & 20.1 & 27.1 & 61.1 & 8.5 & 19.4 & 26.5 \\

\textbf{+Both}  
& 55.8 & 9.8 & 21.4 & 28.6 & 56.0 & 9.6 & 21.0 & 28.5 
 & 55.4 & 10.0 & 22.1 & 28.6  & 56.1 & 9.6 & 21.8 & 28.3  \\ %& 57.0 & 8.9 & 20.2 & 27.4  
 
\midrule

OpenCLIP~\cite{ilharco_gabriel_2021_5143773}  
& 18.9 & 16.9 & 33.3 & 42.0 & 18.9 & 16.9 & 33.3 & 42.0  
& 19.0 &  16.0 & 32.5 & 41.5 & 19.0 &  16.0 & 32.5 & 41.5 \\

\textbf{+Ingredient}  
& 16.0 & 18.0 & 35.5 & 44.2 & 16.0 & 17.5 & 35.0 & 43.8  
& 17.0 & 17.4 & 34.8 & 43.7 & 17.0 & 16.9 & 34.3 & 43.3 \\

\textbf{+Action}  
& 16.0 & 18.2 & 35.5 & 44.2 & 16.3 & 17.6 & 35.0 & 43.6  
& 17.0 & 17.2 & 34.4 & 43.2 & 17.0 & 16.7 & 34.7 & 42.9 \\

\textbf{+Both}  
& 15.4 & 18.4 & 35.8 & 44.5 & 16.0 & 18.0 & 35.1 & 44.1  
& 17.0 & 17.8 & 34.8 & 44.0 & 17.0 & 17.1 & 34.3 & 43.4   \\

\bottomrule
\end{tabular}
}
\label{tab:comparison_both_directions}
\end{table*}

%% file: tables_sup/8_cookpad_region_wise_comparison.tex
\begin{table*}
\centering
\caption{MedR and Recall@\{1,5,10\} results for five cultures (ID: Indonesia, MY: Malaysia, TH: Thailand, VN: Vietnam, IN: India).}

\resizebox{\textwidth}{!}{
\begin{tabular}{@{}l|cccc|cccc|cccc|cccc|cccc@{}}
\toprule
\multirow{2}{*}{} 
& \multicolumn{4}{c|}{\textbf{ID}} 
& \multicolumn{4}{c|}{\textbf{MY}} 
& \multicolumn{4}{c|}{\textbf{TH}} 
& \multicolumn{4}{c|}{\textbf{VN}} 
& \multicolumn{4}{c}{\textbf{IN}} \\ \cmidrule(l){2-21} 
& medR & R@1 & R@5 & R@10 
& medR & R@1 & R@5 & R@10 
& medR & R@1 & R@5 & R@10 
& medR & R@1 & R@5 & R@10 
& medR & R@1 & R@5 & R@10 \\ \midrule

NLLB-SigLIP~\cite{visheratin2023nllb}   & 119.9 & 5.6 & 14.5 & 20.1 & 95.3 & 8.6 & 18.5 & 24.1 & 106.7 & 6.8 & 16.6 & 23.2 & 420.5 & 2.3 & 7.2 & 10.9 & 312.8 & 2.4 & 7.5 &11.0  \\  
\textbf{+Ingredient}  & 115.3 & 5.5 & 14.5 & 20.2 & 86.9 & 8.6 & 17.9 & 24.6 & 83.6 & 7.5 & 17.9 & 24.2 & 426.9 & 2.5 & 7.7 & 10.9 & 301.9 & 2.7 & 8.1 & 12.0 \\
\textbf{+Action}      & 133.5 & 5.7 & 14.6 & 20.1 & 91.4 & 9.0 & 18.6 & 24.6 & 104.1 & 7.7 & 17.7 & 23.4 & 411.4 & 2.5 & 7.2 & 10.4 & 301.1 & 2.3 & 7.2 & 11.1 \\
\textbf{+Both}        & 113.8 & 6.8 & 15.7 & 21.8 & 79.9 & 9.5 & 19.9 & 25.8 & 87.3 & 7.9 & 17.7 & 25.1 & 329.3 & 2.8 & 8.1 & 12.0 & 300.5 & 2.6 & 7.8 & 11.8 \\ \midrule

M-CLIP~\cite{carlsson2022cross}    & 35.5 & 10.1 & 23.1 & 31.7 & 29.6 & 17.6 & 30.7 & 37.7 & 30.9 & 13.1 & 27.4 & 35.1 & 123.7 & 4.2 & 13.2 & 18.3 & 373.5 & 2.1 & 6.3 & 9.3  \\
\textbf{+Ingredient}  & 29.2 & 11.2 & 25.9 & 34.6 & 23.3 & 17.6 & 30.7 & 37.7 & 27.7 & 13.2 & 28.5 & 37.0 & 108.3 & 4.5 & 13.6 & 19.8 & 300.7 & 2.2 & 6.9 & 10.6  \\  
\textbf{+Action}      & 28.6 & 11.5 & 25.2 & 34.1 & 21.9 & 17.8 & 31.4 & 40.0 & 26.4 & 12.8 & 28.4 & 36.8 & 102.2 & 4.9 & 13.8 & 19.8 & 302.6 & 2.3 & 7.4 & 11.2   \\  
\textbf{+Both}        & 28.0 & 11.1 & 25.2 & 34.1 & 23.2 & 18.3 & 32.5 & 40.7 & 24.5 & 13.2 & 29.1 & 38.1 & 99.6 & 4.8 & 14.2 & 20.5 & 276.1 & 2.0 & 6.8 & 10.4 \\ \midrule

OpenCLIP~\cite{ilharco_gabriel_2021_5143773}   & 14.9 & 15.7 & 33.7 & 44.0 & 11.0 & 23.9 & 40.6 & 49.0 & 5.9 & 27.6 & 49.7 & 59.4 & 25.0 & 12.0 & 27.8 & 36.8  & 74.8 & 5.7 & 15.3 & 21.4  \\
\textbf{+Ingredient}  & 14.4 & 16.1 & 35.1 & 45.1 & 9.0 & 25.3 & 43.2  & 51.8 & 5.0 & 28.8 & 51.5 & 60.7 & 20.5 & 13.0 & 30.1 & 38.8  & 64.5 & 6.6 & 17.8 & 24.9  \\ 
\textbf{+Action}      & 14.4 & 16.2 & 34.6 & 44.9 & 9.7 & 25.6 & 43.2 & 51.0 & 5.0 & 28.5 & 51.4 & 61.0 & 20.1 & 13.5 & 31.0 & 39.9  & 60.9 & 6.9 & 17.7 & 24.6 \\
\textbf{+Both}        & 13.8 & 16.5 & 35.4 & 45.8 & 9.9 & 25.8 & 43.1 & 50.8 & 5.0 & 29.0 & 52.0 & 61.5 & 18.9 & 13.4 & 30.9 & 40.8 & 63.9 & 7.1 & 17.9 & 24.2 \\ 
\bottomrule
\end{tabular}
}
\label{tab:supp_cookpad_comparison}
\end{table*}

%% file: tables_sup/9_confusion_matrix.tex
\begin{table}[H]
    \centering
        \caption{Normalized confusion matrix of the culture-predicting classifier.}

    \begin{tabular}{llllll}
\hline
   & Indonesia & Malaysia & Thailand & Vietnam & India \\ \hline
Indonesia &  0.39  &  0.29  &  0.11  &  0.17  &  0.04  \\ \hline
Malaysia & 0.23 & 0.58   &  0.06  &  0.09  &  0.04  \\ \hline
Thailand &  0.06  &  0.06  &  0.66  &  0.20  &  0.02  \\ \hline
Vietnam &  0.05  &  0.02  &  0.20  &  0.70  &   0.03 \\ \hline
India &  0.01  &  0.01  &  0.01  &  0.03  &   0.94 \\ \hline
\end{tabular}
    \label{tab:9_confusion_matrix}
\end{table}

%% file: tables_sup/10_time_comparison.tex
\begin{table}[H]
    \centering
        \caption{Inference speed comparison per query.}

    \begin{tabular}{llll}
\hline
         & None &  Single & Both \\ \hline
H-T~\cite{salvador2021revamping}      &  5.9ms   &  6.7ms & 7.1ms   \\ \hline
TFood~\cite{shukor2022transformer}    &  6.1ms  &  8.8ms  &  10.6ms   \\ \hline
VLPCook~\cite{shukor2022structured}  &  6.6ms  &  11.2ms  &  15.3ms  \\ \hline
OpenCLIP~\cite{ilharco_gabriel_2021_5143773}  &  7ms   &  11ms  &  13ms  \\ \hline
\end{tabular}
    \label{tab:test_time}
\end{table}

%% file: tables_sup/14_cam_visulization.tex
\begin{figure*}
    \centering
    \includegraphics[width=0.75\linewidth]{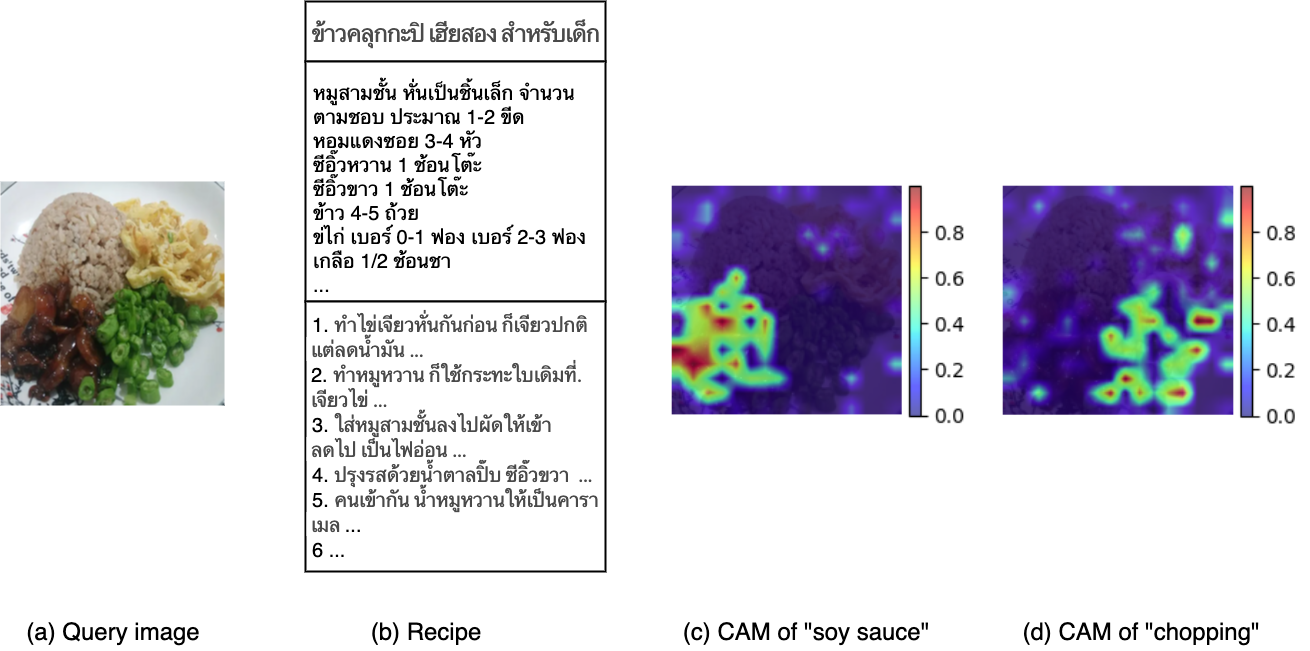}
    \caption{Class activation map (CAM) of ingredient and action prediction: (a) the query image, (b) the corresponding recipe, (c) class activation map of ingredient ``soy sauce'', and (d) class activation map of ingredient ``chopping''.}
    \label{fig:cam_visulization}
\end{figure*}

%% file: tables_sup/13_cam_cookpad.tex
\begin{figure*}
    \centering
    \includegraphics[width=0.65\linewidth]{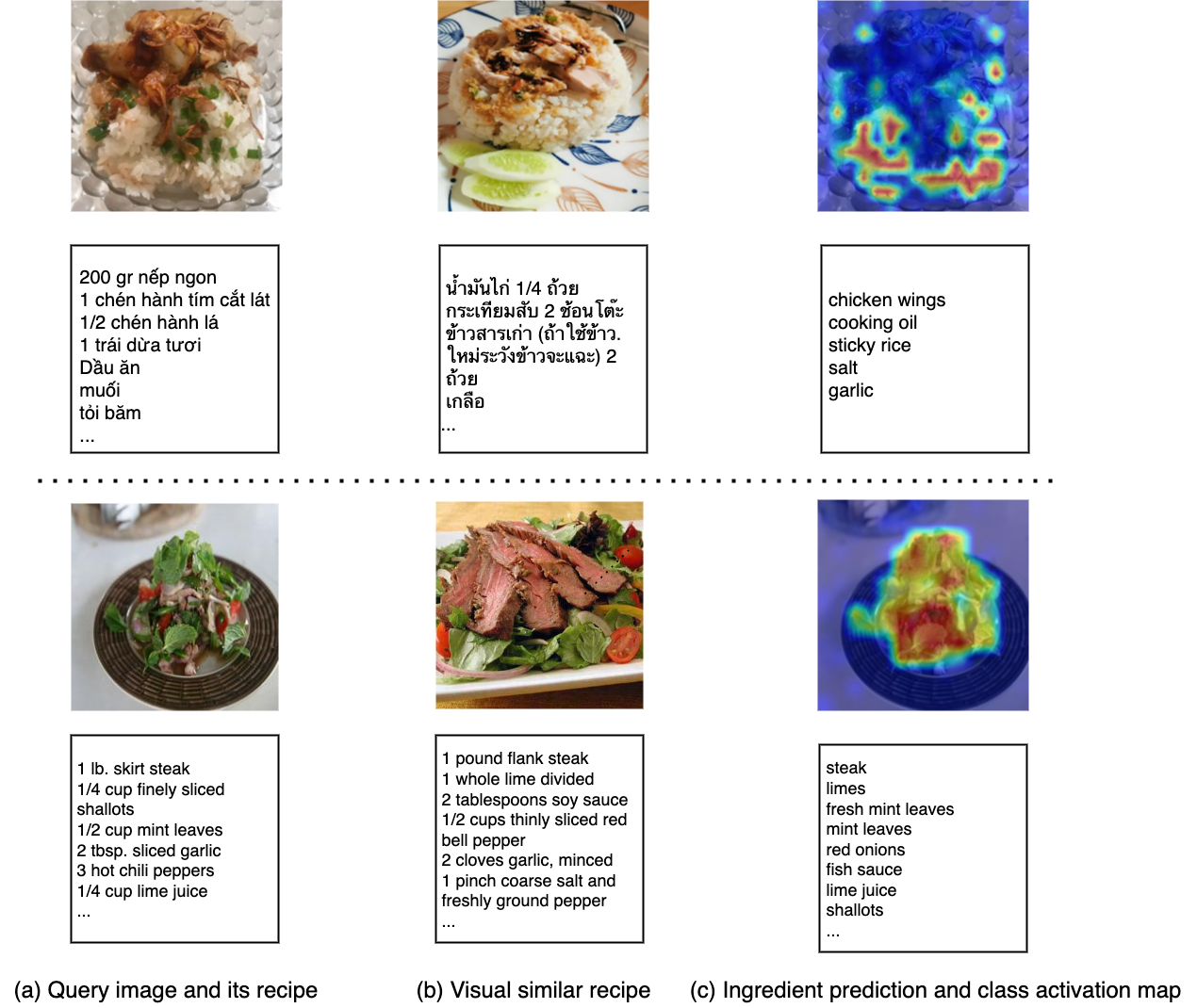}
    \caption{Distinguishing visually similar recipes: (a) the query image and its ingredient composition, (b) a visually similar image and its ingredients, and (c) ingredient prediction with class activation map visualization of ``sticky rice'' (top) and ``lime juice'' (bottom).}
    \label{fig:cam_cookpad}
\end{figure*}

%% file: tables_sup/12_cookpad_error.tex
\begin{figure*}
    \centering
    \includegraphics[width=0.6\linewidth]{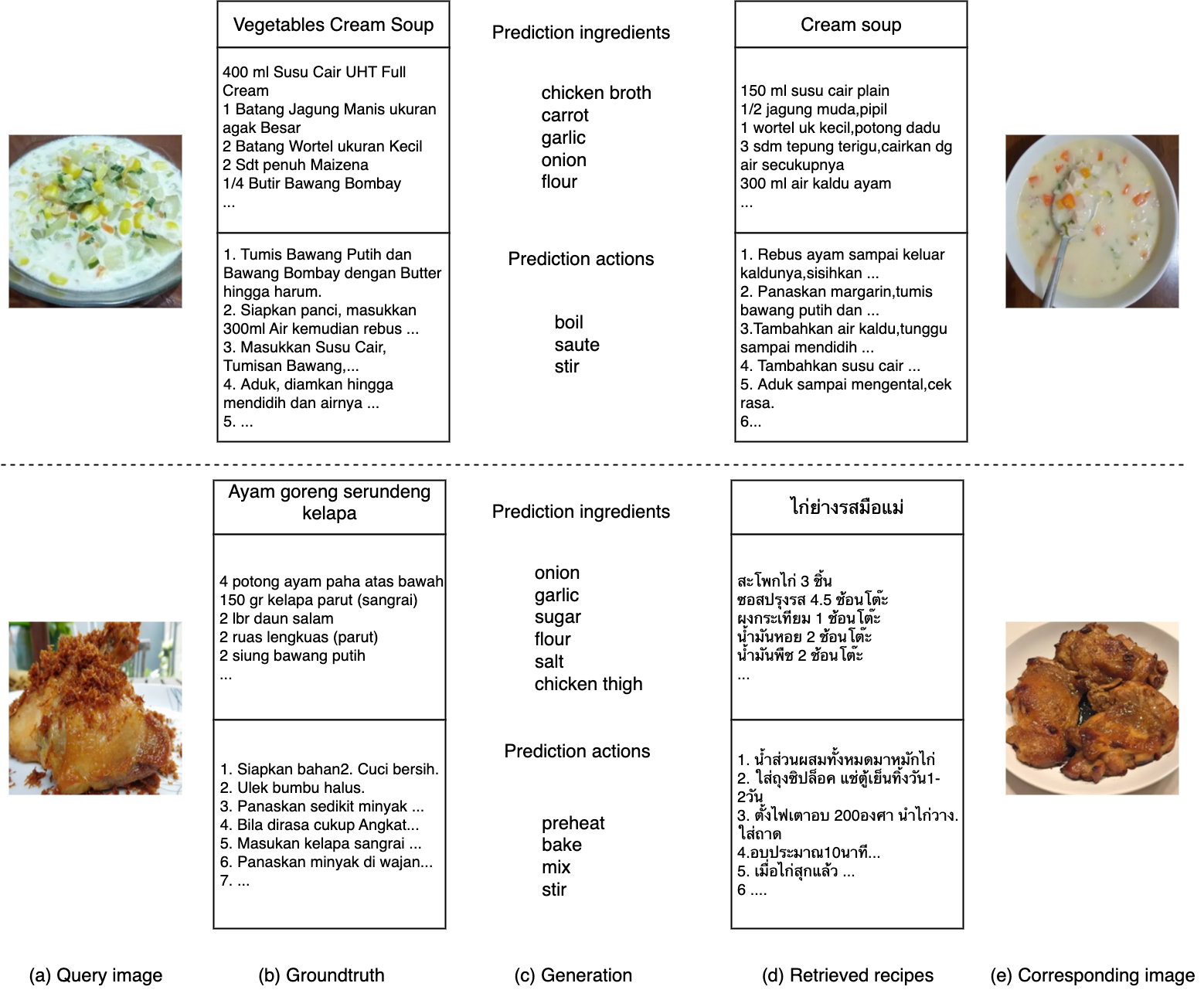}
    \caption{Failure examples in multicultural retrieval: query image (a), the corresponding ground-truth recipe (b), predicted ingredients and actions (c), retrieved recipe and its associated image by debiasing OpenCLIP with both ingredient and action (d) and (e). Common failure cases occur with dishes that are either covered by soup (top) or obscured by toppings (bottom).}
    \label{fig:cookpad_error}
\end{figure*}